\documentclass[lettersize,journal]{IEEEtran}
\usepackage{amsmath,amsfonts}
\usepackage{algorithmic}
\usepackage{algorithm}
\usepackage{array}
\usepackage[caption=false,font=normalsize,labelfont=sf,textfont=sf]{subfig}
\usepackage{textcomp}
\usepackage{stfloats}
\usepackage{url}
\usepackage{verbatim}
\usepackage{graphicx}
\usepackage{cite}
\usepackage{multirow}
\usepackage{color}

\begin{document}

\title{Towards Understanding and Harnessing the Effect of Image Transformation in Adversarial Detection}

\author{Hui Liu, Bo Zhao, Yuefeng Peng, Weidong Li, and Peng Liu~\IEEEmembership{Member,~IEEE,}
        % <-this % stops a space
\thanks{Hui Liu, Bo Zhao, Yuefeng Peng, Weidong Li are with the School of Cyber Science and Engineering, Wuhan University, Wuhan, 430072 China e-mail: zhaobo@whu.edu.cn.}% <-this % stops a space

\thanks{Peng Liu is with the College of Information Sciences and Technology, Pennsylvania State University, PA, 16801 US e-mail: pliu@ist.psu.edu.}}

\markboth{Journal of \LaTeX\ Class Files,~Vol.~14, No.~8, August~2021}%
{Shell \MakeLowercase{\textit{et al.}}: A Sample Article Using IEEEtran.cls for IEEE Journals}


\maketitle

\begin{abstract}
Deep neural networks (DNNs) are threatened by adversarial examples. Adversarial detection, which distinguishes adversarial images from benign images, is fundamental for robust DNN-based services. Image transformation is one of the most effective approaches to detect adversarial examples. During the last few years, a variety of image transformations have been studied and discussed to design reliable adversarial detectors. In this paper, we systematically synthesize the recent progress on adversarial detection via image transformations with a novel classification method. Then, we conduct extensive experiments to test the detection performance of image transformations against state-of-the-art adversarial attacks. Furthermore, we reveal that each individual transformation is not capable of detecting adversarial examples in a robust way, and propose a DNN-based approach referred to as \emph{AdvJudge}, which combines scores of 9 image transformations. Without knowing which individual scores are misleading or not misleading, \emph{AdvJudge} can make the right judgment, and achieve a significant improvement in detection rate. Finally, we utilize an explainable AI tool to show the contribution of each image transformation to adversarial detection. Experimental results show that the contribution of image transformations to adversarial detection is significantly different, the combination of them can significantly improve the generic detection ability against state-of-the-art adversarial attacks. % \emph{AdvJudge} is a more effective adversarial detector than those based on an individual image transformation. 
\end{abstract}


\section{Introduction}

\IEEEPARstart{A}{dversarial} attacks~\cite{Carlini,Goodfellow,Kurakin,DuanY,ChenS}, which can generate images with small but maliciously-crafted perturbations to change the output of state-of-art classifiers, are still problematic for deep neural networks (DNNs)~\cite{HeK, HuangG, Karen, HuJ}. These perturbed images are referred to as adversarial examples~\cite{Szegedy}. Adversarial attacks have proved to be a serious threat to DNN-based services in the physical world~\cite{Sharif,Lovisotto,RenH,WangJ}. For example, attackers who wear a pair of well-designed eyeglass frames can evade detection by facial recognition systems~\cite{Sharif}. Attackers can project specifically crafted adversarial perturbations onto real-world objects, transforming them into adversarial examples~\cite{Lovisotto}.

The existence of adversarial examples raises great public concern and motivates research for an proactive defense. These proactive defenses, e.g., adversarial training~\cite{Ding2020MMA,HoJ,ZhangJ}, gradient masking~\cite{Hyungyu,LiY,AprilPyone,Athalye}, etc., can harden DNN systems with small accuracy loss. The adversarial training employs as many adversarial examples as possible during training process as a kind of regularization. The gradient masking defends against adversarial attacks by not providing the attacker with useful gradients of the DNN model. However, it is shown in literature~\cite{TanJ,Tramer,BaiT} that hardened DNNs can still be evaded.

Due to the limitations of proactive defenses, reactive defenses, i.e., adversarial detection~\cite{Lust,LuoW,KianiS} has attracted increasing interest of the research community. Unlike proactive defenses, adversarial detection only determines whether an input is benign or adversarial, without the need to identify the ground-truth label of an input. 

Image transformation~\cite{Aldahdooh,Nesti,XuW,LiuH,Graese,Bahat,TianS,Mekala,Guesmi,ShouW}, which measures the disagreement of the DNN model in predicting the input and its transformed version, is one of the most effective approaches to detect adversarial examples. The main challenge of detecting adversarial examples via image transformation is to design appropriate transformation approaches, which should have the following two properties: (i) the DNN's prediction on the adversarial example is significantly changed, and (ii) the DNN's prediction on the benign example is unaffected. These two properties can be employed to identify the authenticity of the test image.

This paper systematically synthesizes (the knowledge gained through) the recent progress on adversarial detection via image transformations with a novel classification. Notably, we test the existing adversarial detectors based on image transformations against state-of-the-art adversarial attacks, and reveal a fundamental limitation of image transformation based detectors: no individual kind of image transformation has generic (i.e., can detect all kinds of adversarial attacks) detection ability. This key observation motivates us to answer a new research question: Is it possible to combine the strengths of each individual image transformation while avoiding their weaknesses? To answer this question, we propose to learn (the truth) from the individual detector scores without any prior knowledge about which individual scores are misleading. We design a DNN-based \emph{AdvJudge}, and conclude that adversarial examples could be detected in a robust way by combining multiple transformations. Furthermore, we utilize an explainable AI tool to understand how our \emph{AdvJudge} is working and it's decision.

In summary, this paper makes the following contributions:
\begin{itemize}
\item To the best of our knowledge, we are the first to systematically review and test adversarial detection via image transformations with a novel classification. 

\item We reveal that an individual image transformation has no generic ability to detect adversarial examples. Thus, we combine scores of 9 image transformations and design a DNN-based detector referred to as \emph{AdvJudge}. Experimental results show that \emph{AdvJudge} significantly outperforms all individual transformation detectors.

\item Beyond considering the accuracy of the \emph{AdvJudge}, we leverage an explainable AI tool to reveal which image transformations are more influential and how \emph{AdvJudge} realizes its detection. 
\end{itemize}

The remainder of this paper is organized as follows. we introduce the related work in Section \uppercase\expandafter{\romannumeral2}. Section \uppercase\expandafter{\romannumeral3} presents preliminaries on the adversarial attack and detection. In Section \uppercase\expandafter{\romannumeral4}, we present a novel classification of image transformations for adversarial detection in the baseline architecture. Section \uppercase\expandafter{\romannumeral5} evaluates the performance of each of the image transformations in adversarial detection. In Section \uppercase\expandafter{\romannumeral6}, we design and test the proposed detector \emph{AdvJudge} that combines 9 image transformations. Finally, we conclude the paper in Section \uppercase\expandafter{\romannumeral7}.

\section{Related Work}
Image transformation is one of the significantly important and effective methods for adversarial detection~\cite{Aldahdooh,Nesti,XuW,LiuH,Graese,Bahat,TianS,Mekala,Guesmi}. This kind of method is based on a key observation: adversarial examples are more sensitive to image transformations comparing with benign examples. This observation motivates researchers to design adversarial detectors by measuring the prediction inconsistency in an unknown input and its transformed versions.

A large number of state-of-the-art adversarial detectors based on image transformation have been proposed in recent years~\cite{Aldahdooh}. Xu et al.~\cite{XuW} observed that the inputs provided unnecessarily large feature spaces for an adversary to generate adversarial examples. They introduced feature squeezing as a technique for adversarial detection, and proposed two instances of squeezing: reducing the color depth of images, and using smoothing to reduce the variation among pixels. If the difference between the model's prediction on the unknown input and its prediction on the squeezed input exceeds a threshold level, this input is considered to be adversarial. Liu et.al~\cite{LiuH} attributed the imperceptibility of adversarial examples to recessive features. They designed a feature-filter based on discrete cosine transform to detect imperceptible adversarial examples. Feature-filter identified adversarial examples by comparing only the prediction labels of the DNN on the input and its filtered version. Their experiments illustrated high performance was achieved when detecting C\&W adversarial examples. Kantaros et al.~\cite{Kantaros} observed that adversarial examples are sensitive to lossy compression transformations. As a transformation, they proposed VisionGuard based on lossy compression algorithms (e.g., JPEG). The experiments showed that the VisionGuard could detect large-scale adversarial examples in real time. Meng et.al~\cite{MagNet} argued that an adversarial example was misclassified for two reasons. The adversarial example is either far from nor close to the boundary of the manifold of the task. Therefore, they proposed a two-pronged detection methods MagNet against adversarial examples. The first one combines the reconstruction error and the probability divergence to measures the distance between the unknown example and the manifold of benign examples. The second way measures the distances between the predictions of unknown inputs and their denoised versions. The experiments showed that MagNet was effective against the state-of-the-art adversarial attacks in blackbox and graybox scenarios.

These well-designed detection methods verify that image transformation can effectively detect adversarial examples without accuracy loss. However, the complicated design obscures the effect of each image transformation in adversarial detection. How the basic image transformation affects the adversarial detector remains to be explored.

A few studies have focused on the role of basic image transformation in adversarial detection~\cite{Nesti,TianS,Mekala}. Nesti et.al~\cite{Nesti} conducted an extensive experimental evaluation on detection capabilities of image transformations, including translation, rotation, shear, average blur, etc. They tested and reported the performance of these transformations based on the baseline and voting architectures. Tian et.al~\cite{TianS} applied a set of rotation operations on an unknown image to generate several transformed images, and then used the classification results of these transformed images as features to predict if this image is adversarial. The Experiments showed that their method was effective against the C\&W attack in both oblivious attacks and white-box attacks. Based on metamorphic testing principles, Mekala et.al~\cite{Mekala} applied affine transformations to design their detection architecture. This architecture detected adversarial examples by comparing the classification of an image with that of its affine transformed version. These affine transformation functions linearly modifies the spatial orientation of the images, e.g. translation, scale, rotation and shear. They claimed that the four affine transformations achieved high accuracy ranges bordering around 90\%, and the shear and translation were the most effective in detecting adversarial examples.

Existing studies have proved that the basic image transformations can help to detect adversarial examples. The recent advances have resulted in a variety of new image transformations for adversarial detection, which no doubt deserves a comprehensive review. Beyond verifying the detection performance of image transformations, there are some questions to understand which transformations are important for adversarial detection and how the adversarial detector is working.

\section{Preliminaries}
\subsection{Convolutional Neural Network}
Convolutional neural networks (CNNs)~\cite{HeK, HuangG, Karen, HuJ} are the most widely used neural network in recent neural network architectures, which perform incredible success in computer vision tasks. CNNs can be formalized as a function $f(x,\theta) = y$ that accepts an $x \in \mathbb{R}^n$ and produces an output $y \in \mathbb{R}^m$. They are usually constructed by multiple convolutional layers and nonlinear activation functions, e.g., sigmoid, ReLu, and tanh.

In this paper, we test the performance of detectors based on image transformations over the well-trained Inception V3~\cite{Karen}.

\subsection{Adversarial Attack}

An adversarial attack~\cite{Carlini,Goodfellow,Kurakin,DuanY,ChenS} is a unique form of attack in deep learning. Attackers mislead state-of-the-art DNN-based classifiers into making erroneous predictions by adding small perturbations to inputs. Most studies follow the formulation proposed by Szegedy et al. in ~\cite{Szegedy}, where the generation of adversarial attacks is formulated as a minimization problem with the perturbation magnitude as the main constraint. That is, 
\begin{equation}
\begin{aligned}
{\rm min} \quad  &\lVert \delta \rVert _p \\
s.t. \quad & f(x) \neq f(x') \\
&x'=x +\delta \in D\\
\end{aligned}
\label{eq2}
\end{equation}
where $p$ norms include $L_0$, $L_2$ and $L_{\infty}$ to measure the magnitude of the perturbation, $\delta$ denotes the perturbation matrix that is added to the benign example \emph{x} for synthesizing the adversarial example $x^\prime$, which remains in the benign domain $D$. Function $f(x) = y$ is a well-trained DNN model, which accepts an $x \in \mathbb{R}^n$ and produces an output $y \in \mathbb{R}^m$. 
\iffalse
Convolutional neural networks (CNNs) are the most widely used neural network architectures and have achieved incredible success in computer vision tasks. They are usually constructed by multiple convolutional layers and nonlinear activation functions, e.g., sigmoid, ReLu, and tanh.
\fi

\subsubsection{Carlini/Wagner Attacks}
Carlini/Wagner (C\&W) attacks~\cite{Carlini} proposed three gradient -based algorithms to achieve imperceptible adversarial perturbations, which defeat defensive distillation and are more effective than some state-of-the-art approaches to attack potential defenses. Among three attacks, $L_2$ norm performs the most powerful attack ability. C\&W attacks with $L_2$ norm could be formulated as follows,
\begin{equation}
\begin{aligned}
{\rm min} \quad & \lVert \delta \rVert _2 + c \cdot Loss(x') \\
s.t. \quad & x'=x+\delta \in D\\
\end{aligned}
\label{cw1}
\end{equation}
where $c$ is a hyper-parameter, and \emph{Loss} is defined as
\begin{equation}
Loss(x')={\rm max}({\rm max}\{Z(x')_i : i \neq tg\} - Z(x')_{tg}, -k)
\label{cw2}
\end{equation}
where $Z(x)$ is the output of the network before the softmax layer, and a hyper-parameter $k$ encourages the attack to search for an adversarial example $x^\prime$ that will be classified as label $tg$ with high confidence.

\subsubsection{Fast Gradient Sign Method}
Goodfellow et al.~\cite{Goodfellow} explained the existence of adversarial examples for DNN's linear nature, and proposed fast gradient sign method (FGSM) to efficiently generate adversarial perturbations. For untargeted attack, the adversarial example is generated by adding a gradient vector of the loss function as follow.
\begin{equation}
x'=x+\epsilon \cdot \operatorname{sign}\left(\nabla_{x} J(f(x), y)\right)
\label{fgsm}
\end{equation}
where $\epsilon$ is a perturbation level to control the amount of perturbations. $J(\cdot,\cdot)$ is loss function, $y$ is the ground-truth label for $x$.

\subsubsection{Basic Iterative Mthod}
Basic iterative method (BIM)~\cite{Kurakin} is an extension of FGSM attack. It applies FGSM multiple times with small step size, and clips pixel values of intermediate results after each step to ensure that they are in an $\epsilon$-neighbourhood of the original image $x$.
\begin{equation}
\begin{aligned}
x'_{0}=x, x'_{m+1}=\operatorname{Clip}_{x, \epsilon}\left\{x'_{m}+\alpha \operatorname{sign}\left(\nabla_{x} J\left(x'_{m}, y\right)\right)\right\}
\end{aligned}
\label{bim}
\end{equation}
where $m$ is the number of iterations, $\alpha$ indicates the magnitude of pixel values changed on each step, $\operatorname{Clip}_{x, \epsilon}(\cdot)$ performs per-pixel clipping to keep the result in the $L_{\infty}$ $\epsilon$-neighbourhood of $x$. 

\subsection{Prediction Distance}
The DNN prediction is usually a vector, which represents the probability distribution how likely an input sample is to belong to each possible class. Existing approaches usually utilize Kullback-Leibler divergence ($D_{KL}$)~\cite{Bahat} to measure the distance between two probability distribution values. The definition of $D_{KL}$ is formalized as Equation ~\ref{kl}.
\begin{equation}
D_{KL}(Z(x)||Z(x_t)) = \sum Z(x) log \frac{Z(x)}{Z(x_t)}
\label{kl}
\end{equation}
where $D_{KL}$ is a measure of how a prediction distribution $Z(x)$ of test image $x$ is different from $Z(x_t)$ of its transformed version $x_t$. The $D_{KL}$ value is the expectation of the logarithmic difference between the probabilities $Z(x)$ and $Z(x_t)$. In the special case, a $D_{KL}$ value of 0 indicates that the two probability distributions have identical quantities of information.

\section{Image Transformation for Adversarial Detection}
\subsection{Baseline Detection Architecture}
\begin{figure}[htb]
\centering
\includegraphics[width=0.6\columnwidth]{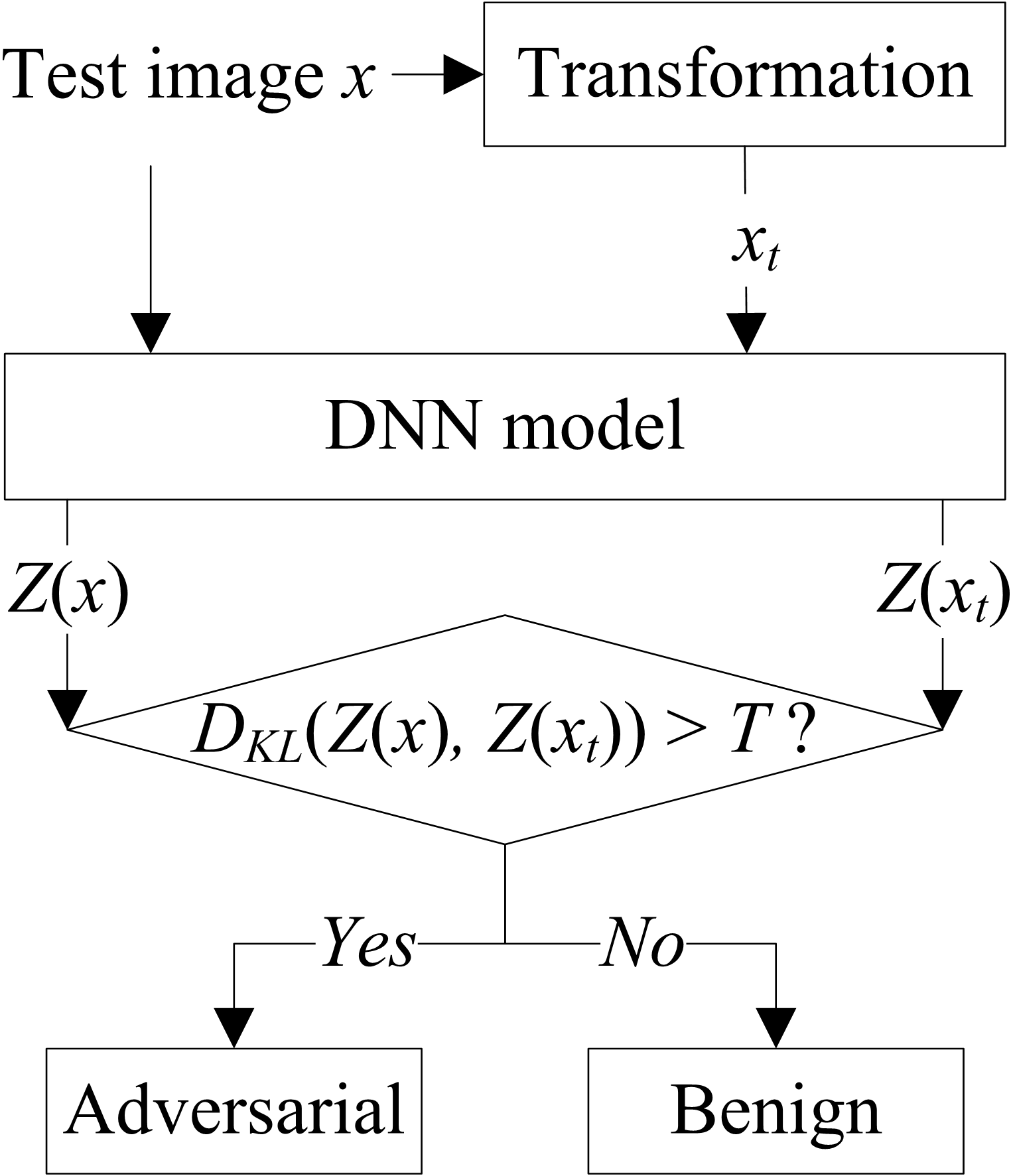}
\caption{Baseline architecture to detect adversarial examples via image transformations.}
\label{archi}
\end{figure}

Adversarial examples yield significantly different outputs of the classifier under certain image transformations. In contrast, benign examples tend to yield robust and consistent classifications under these transformations. Given a test image \emph{x} whose authenticity is unknown, an image transformation is performed to get its transformed version \emph{$x_t$}. The test image \emph{x} and its transformed version \emph{$x_t$} are input to the DNN model. Then, this model produces two predictions, \emph{Z(x)} and \emph{Z($x_t$)}. If the two predictions are far apart from each other, the test image is considered to be adversarial; otherwise, it is considered benign. Figure~\ref{archi} shows the baseline architecture to detect adversarial examples via image transformations.

Adversarial detection via image transformations has the following advantages: (i) they do not modify the architectures or parameters of the neural network and do not result in accuracy loss, (ii) they are independent of neural networks, and complementary to other defenses and detections, and (iii) they are simple, inexpensive, and so can be performed at runtime.

\subsection{Taxonomy}
\begin{figure}[htb]
\centering
\includegraphics[width=0.46\columnwidth]{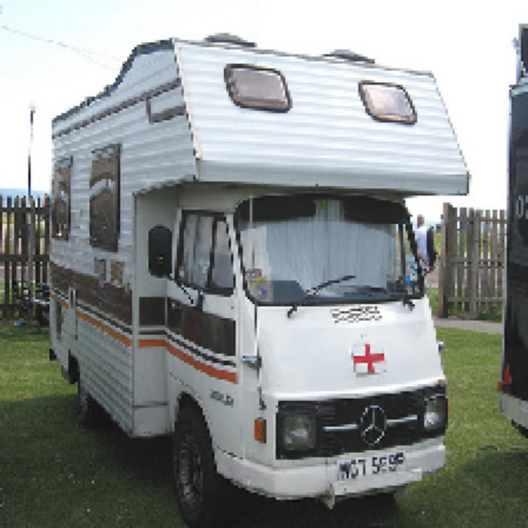}
\caption{An image to be transformed.}
\label{testImage}
\end{figure}

In this subsection, we classify image transformations in adversarial detection into two dimensions: pixel modification, and topological transformation. To visualize image transformations, an image is selected as a transformed object in Figure~\ref{testImage}.

\subsubsection{Pixel Modification}
\begin{figure}[htb]
\centering
\subfloat[Gaussian noise]{
    \includegraphics[width=0.4\columnwidth]{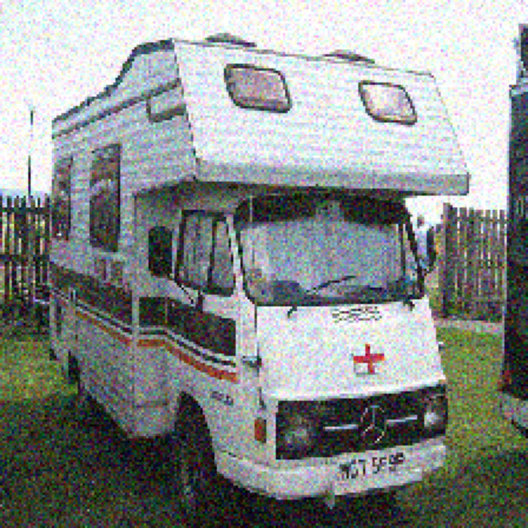}
}\quad
\subfloat[Maximum smoothing]{
    \includegraphics[width=0.4\columnwidth]{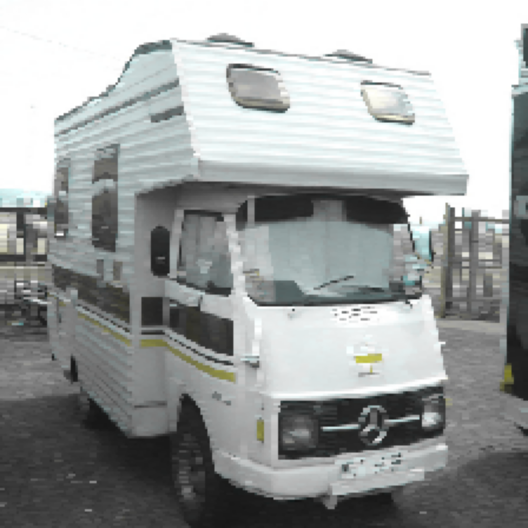}
}  \\
\subfloat[Bit-depth reduction]{
    \includegraphics[width=0.4\columnwidth]{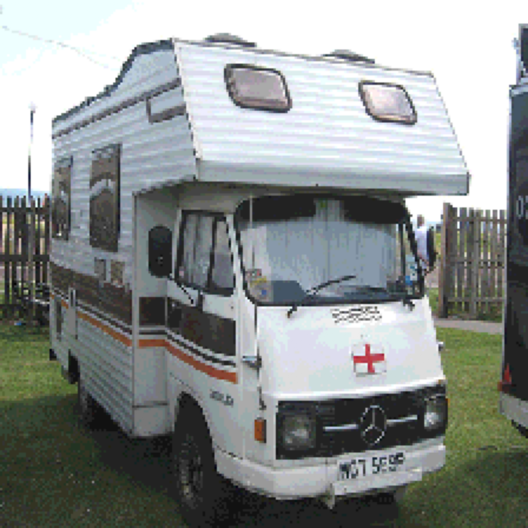}
}\quad
\subfloat[Feature-filter]{
    \includegraphics[width=0.4\columnwidth]{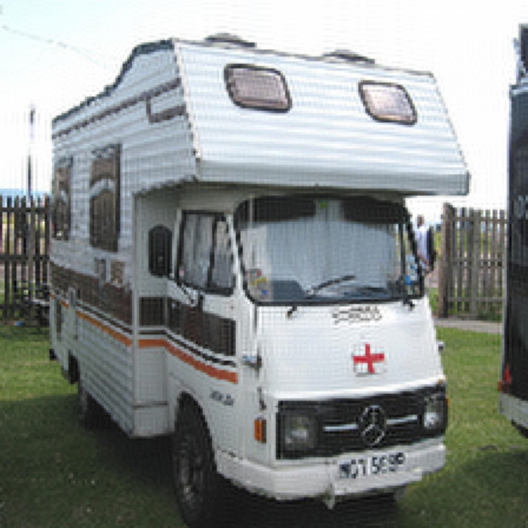}
}
\caption{Visualization of pixel transformation.}
\label{pixelImages}
\end{figure}

Pixel modification is a kind of transformation that modifies the local or whole pixel values in an image. It can be divided into four groups: additive noise, smoothing, bit-depth reduction, and frequency-domain transformation. The transformed images via pixel modification are shown in Figure~\ref{pixelImages}.

(1) Additive noise: Image noise is a random variation of pixel values in an image. The most frequently occurring noises include Gaussian noise, Poisson noise, salt\&pepper noise, etc. Gaussian noise is a statistical additive noise, which has a probability density function equal to that of the Gaussian distribution. Poisson noise is a type of random mathematical object that consists of points randomly located on an image. Salt\&pepper noise, also known as impulse noise, is caused by sharp and sudden disturbances in the image signal. It presents itself as sparsely occurring white and black pixels. Nesti et.al claimed that adversarial examples that are robust to additive Gaussian noise may exist~\cite{Nesti}.

(2) Smoothing: Smoothing is a group of techniques widely used in image processing for reducing image noise. The pixel values are modified so individual pixel values higher than the adjacent points are reduced, and pixel values that are lower than the adjacent points are increased leading to a smoother image. By selecting different mechanisms in weighting the neighboring pixels, the image smoothing method can be divided into the maximum, median, uniform, and Gaussian filters, etc~\cite{XuW}. The maximum filter runs a sliding window over each pixel of the image, where the center pixel is replaced by the maximum value of the neighboring pixels within the window.

(3) Bit-depth reduction: Bit-depth reduction aims to reduce the information capacity of the image representation~\cite{XuW}. This approach is driven by the observation that the feature input spaces are often unnecessary large, and the vast input space provides extensive opportunities for an adversary to generate adversarial examples. Given an input $x \in [0, 1]$, to reduce $x$ to $i$-bit depth ($1 \leq i \leq 7$), the input $x$ is multiplied by $2^i - 1$ and then the result is rounded to an integer. Next the integer is scaled back to [0, 1] by dividing $2^i - 1$. The bit depth is reduced from 8 bits to $i$-bit with the integer-rounding operation.

(4) Frequency-domain: Many adversarial attacks attempt to modify pixels in ways that are usually imperceptible. High frequency patterns play a key role in the generation of adversarial examples~\cite{ZhouY}. Detection strategies based on the frequency-domain usually discard the high-frequency information of the image to ``compress away" such adversarial perturbations. In particular, JPEG compression and its variants in literature~\cite{Kantaros,LiuH} are shown to be an effective countermeasure to imperceptible adversarial examples.

\subsubsection{Topological Transformation}
\begin{figure}[htb]
\centering
\subfloat[Translation]{
    \includegraphics[width=0.4\columnwidth]{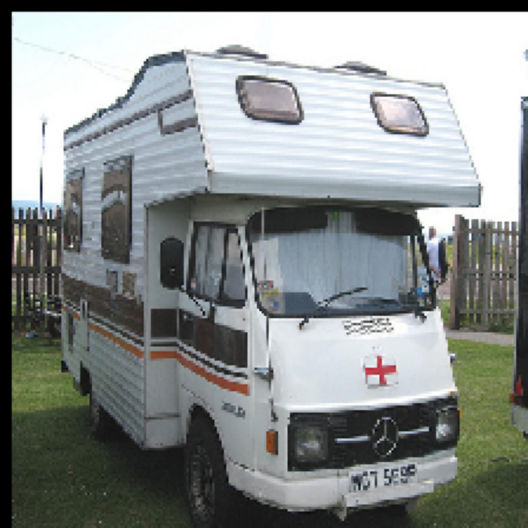}
}  \\
\subfloat[Flip]{
    \includegraphics[width=0.4\columnwidth]{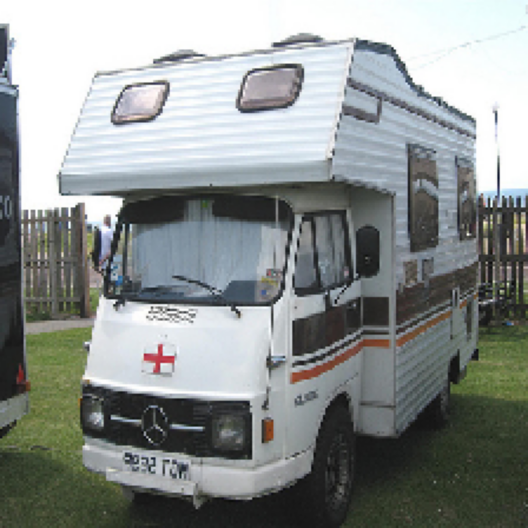}
}\quad
\subfloat[Rotation]{
    \includegraphics[width=0.4\columnwidth]{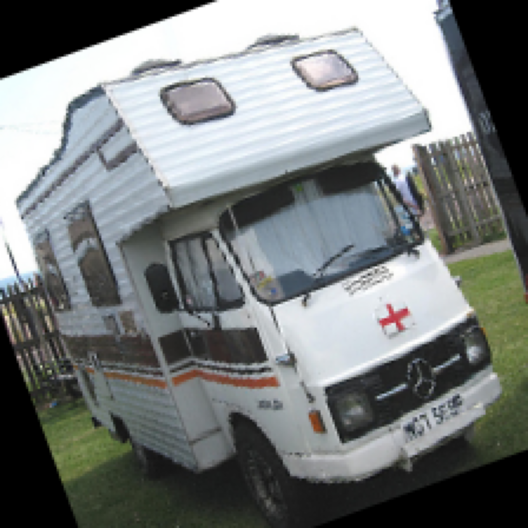}
}  \\
\subfloat[Shear]{
    \includegraphics[width=0.4\columnwidth]{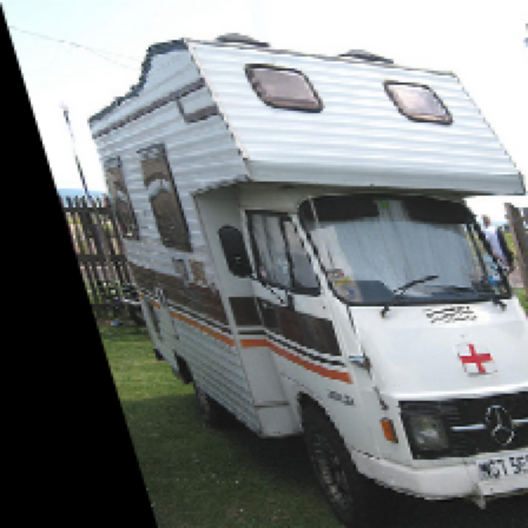}
}\quad
\subfloat[Scale]{
    \includegraphics[width=0.4\columnwidth]{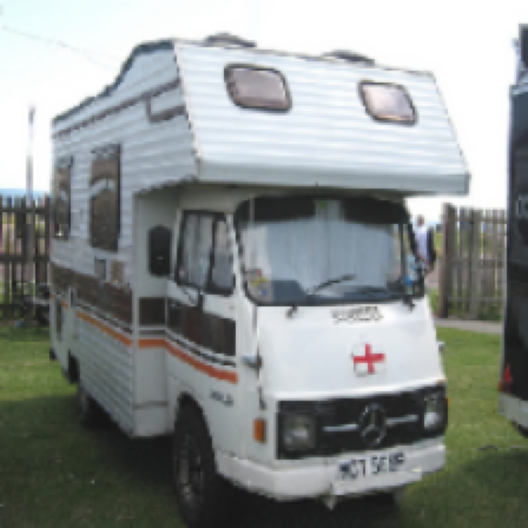}
}
\caption{Visualization of topological transformation.}
\label{topologicalImages}
\end{figure}

Topological transformation includes an affine transformation that preserves both the collinearity relation between points and ratios of distances along the line, and aims to change the pixel coordinates in an image to obtain a transformed version. The basic topological transformations include translation, flip, rotation, shear and scale. The transformed images via pixel modification are shown in Figure~\ref{topologicalImages}.

(1) Translation: Translation is a simple topological transformation, which moves all of the pixels in the same direction. Translation can be formulated as $ v' = v + \sigma $, where $v = [\mu, \nu]^T$ and $v' = [\mu', \nu']^T$ represent the coordinates of a pixel in the original and transformed images, and $ \sigma = [\alpha, \beta]^T $ are horizontal and vertical translation parameters~\cite{Graese}.

(2) Flip: The flipped image is a reflected duplication of an object that appears almost identical but is reversed in the direction perpendicular to the mirror surface. Image flip can be formulated as $v' = T v$, where $T \in \mathbb{R}^{2 \times 2}$ is a warping matrix, e.g., $T = \left[ \begin{matrix} -1 & 0 \\ 0 & 1 \end{matrix} \right] $ in the horizontal flip. The effectiveness of the horizontal flip has been proved against C\&W attacks under the known detector scenario~\cite{Bahat}.

(3) Rotation: Rotation is the linear transformation of space around a point of origin. Rotation has a formalized representation, e.g., $v' = T v$ similar to flip, where $T = \left[ \begin{matrix} cos \theta & -sin \theta \\ sin \theta & cos \theta \end{matrix} \right] $, $\theta$ is the rotation angle~\cite{TianS,Thang}. The authors claimed in literature~\cite{TianS} that their rotation-based detector achieved an accuracy of nearly 99\% against C\&W attacks in the white-box setting.

(4) Shear: Image shear is a linear transformation of each pixel to a fixed direction proportional to that of parallel lines in the plane. The horizontal shear can be formulated as $v' = T v$, where $T = \left[ \begin{matrix} 1 & a \\ 0 & 1 \end{matrix} \right] $, and $a$ is a shear parameter. Experiments in literature~\cite{Mekala} demonstrated that horizontal shear was effective for adversarial detection. 

(5) Scale: Image scaling is a linear transformation that geometrically scales up or down the image dimensions. Scale can be expressed by the $T = \left[ \begin{matrix} s & 0 \\ 0 & s \end{matrix} \right] $. The scale parameter $s > 1$ indicates the zoom-in transformation while $s < 1$ indicates zoom-out. The zoom-in transformation yields an excellent ability to detect untargeted C\&W attacks.

%------------------------------------------------------------------------
\section{Performance of Image Transformations}
\subsection{Experimental Setup}
We randomly choose 1,800 images from the ImageNet dataset~\cite{ImageNet} as benign images, and respectively yield 600 adversarial examples against the Inception V3 model by C\&W~\cite{Carlini}, FGSM~\cite{Goodfellow}, and BIM attacks~\cite{Kurakin}. Benign examples are divided into 1,440 training examples and 360 test examples. Each type of adversarial example includes 480 training examples and 120 test examples. For C\&W attacks, the target class is the next class. The perturbation level $\epsilon$ is set to 0.1 for both FGSM and BIM attacks. Note that the low perturbation level strictly limits the magnitude of adversarial perturbations, making adversarial detection more difficult. Pixel modifications and topological transformations are executed separately on benign examples and adversarial examples. Experimental results show the contribution of each image transformation to adversarial detection.
%-------------------------------------------------------------------------

\subsection{Threshold Setting}
As shown in Figure~\ref{archi}, the appropriate threshold $T$ is very important for the detection accuracy based on the baseline architecture. For this test, training examples are utilized to determine the appropriate threshold, and test examples are utilized to test the detection accuracy under this threshold.

It is time-consuming to determine an appropriate threshold $T$ by a global search. To reduce the complexity, we calculate the $D_{KL}$ value between training images and their transformed versions as the candidate set of the threshold. By iterating through the candidate set, we identify a threshold that maximizes the product of the TPR (true positive rate) and TNR (true negative rate) in training images. Under this threshold, we test the detection ability of the detector based on each individual image transformation against benign examples and three types of adversarial examples.

%-------------------------------------------------------------------------
\subsection{Pixel Modification Tests}
\begin{figure*}[htb]
\centering
\subfloat[Gaussian noise]{
    \includegraphics[width=0.64\columnwidth]{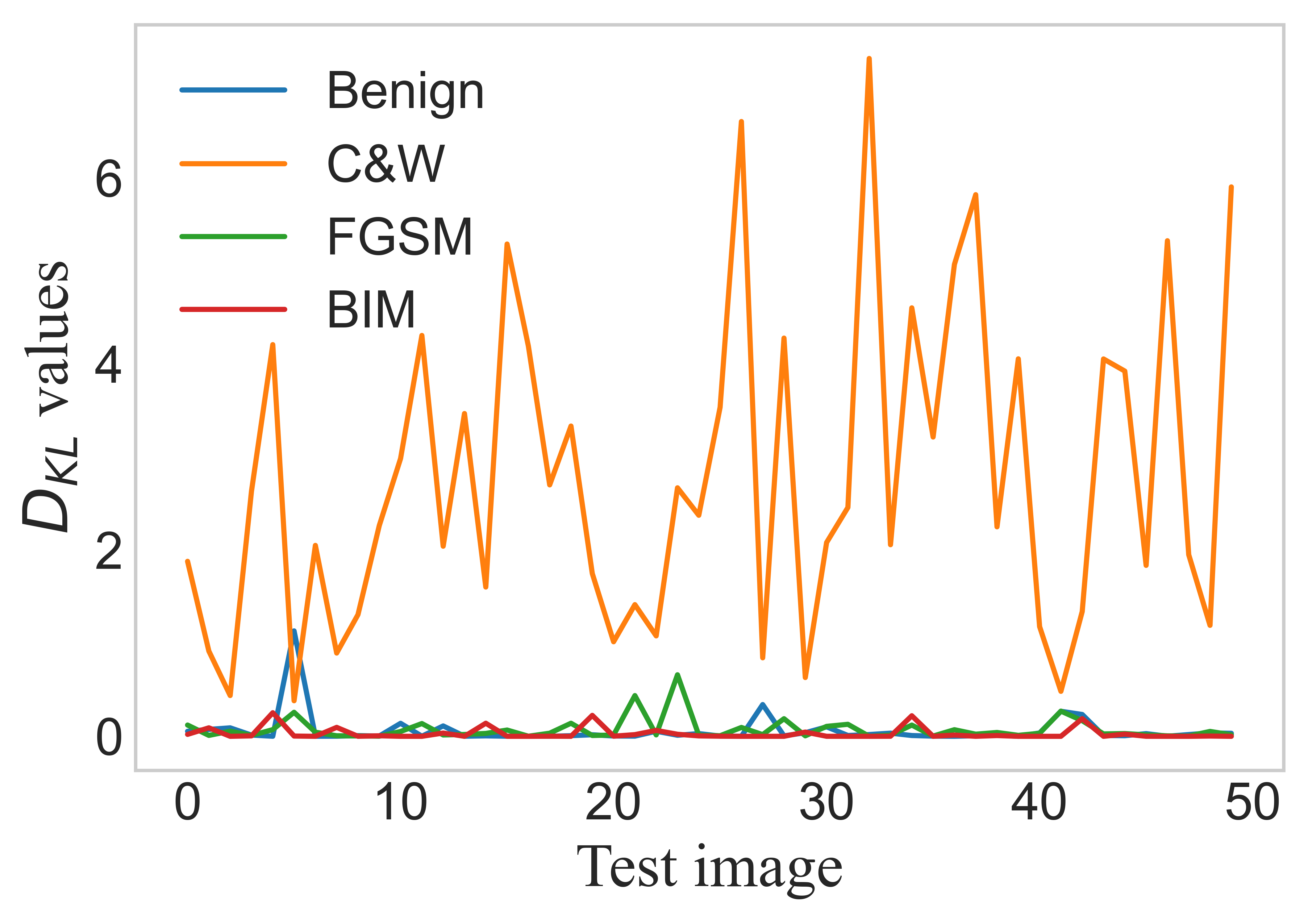}
}
\subfloat[Maximum smoothing]{
    \includegraphics[width=0.679\columnwidth]{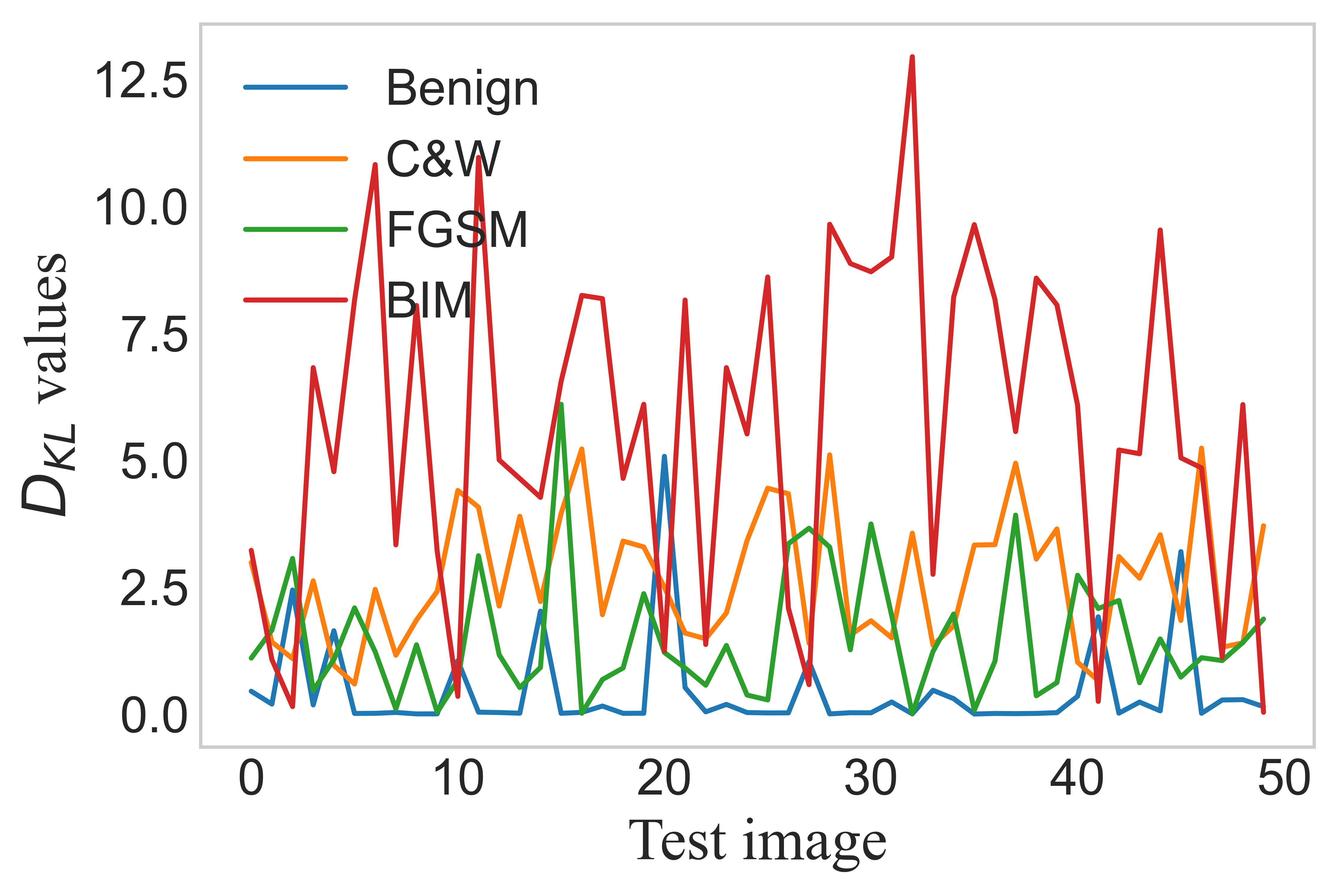}
}  \\
\subfloat[Bit-depth reduction]{
    \includegraphics[width=0.64\columnwidth]{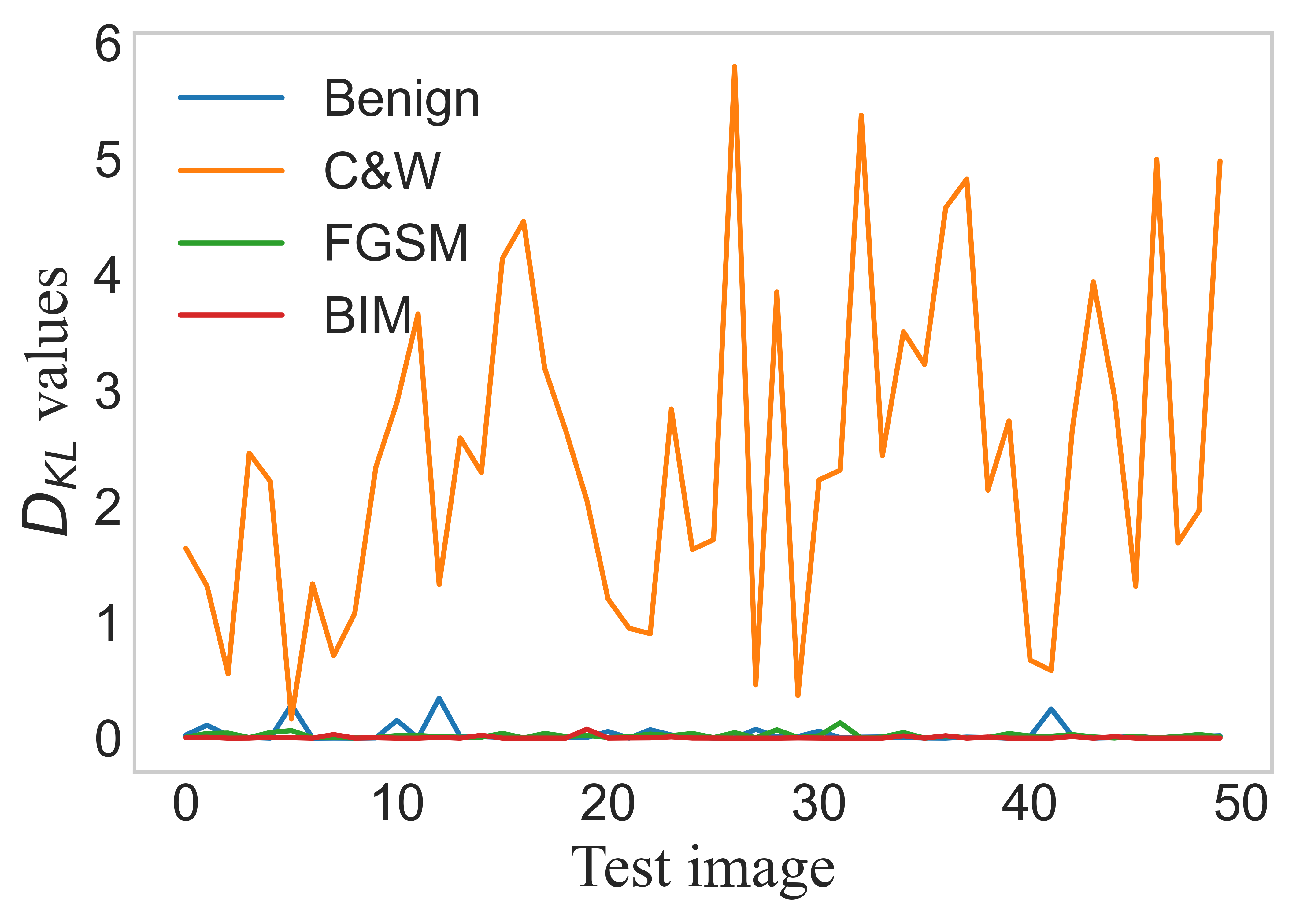}
}
\subfloat[Feature-filter]{
    \includegraphics[width=0.64\columnwidth]{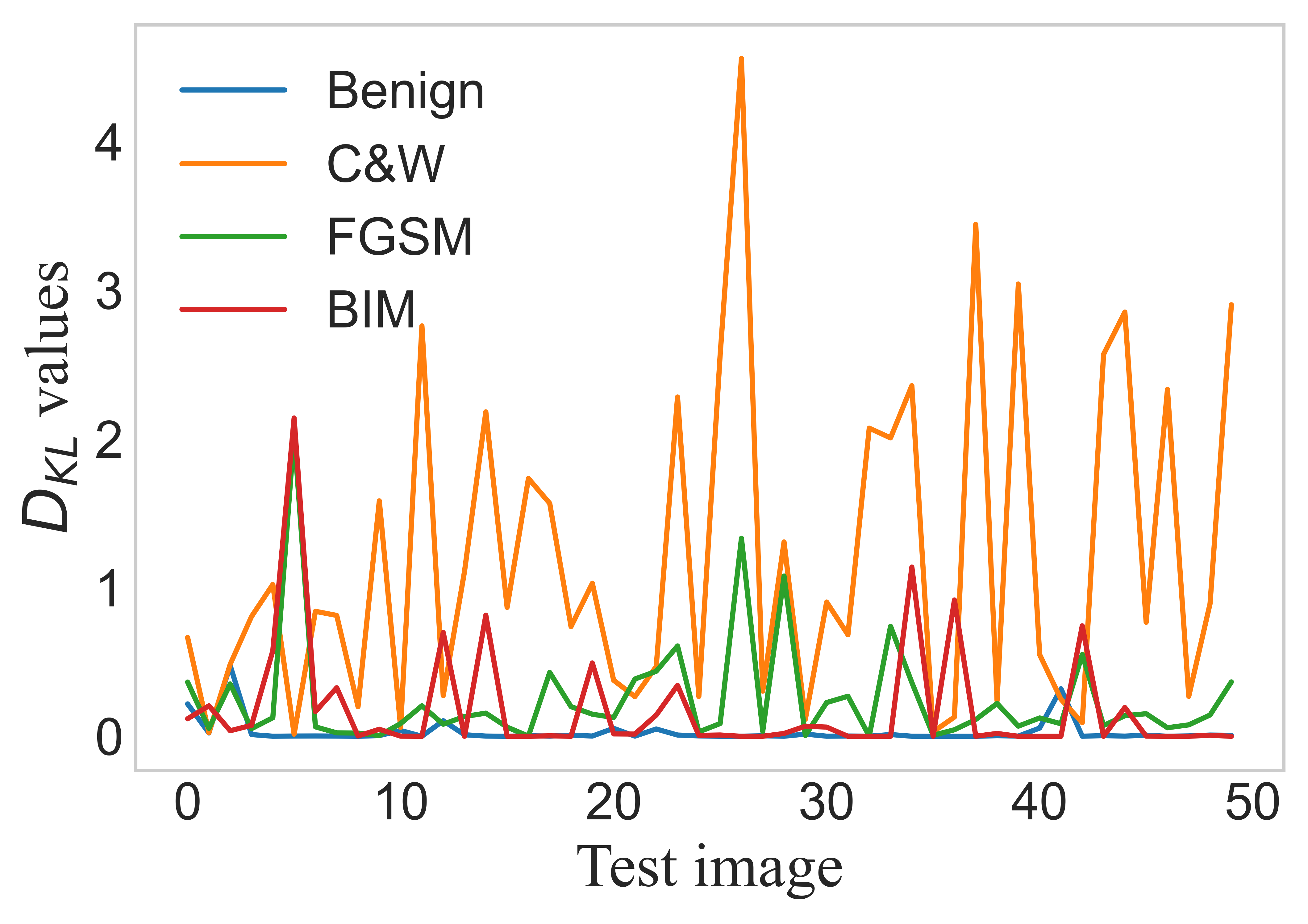}
}
\caption{Distributions of the $D_{KL}$ values via pixel modification. The $D_{KL}$ values of benign images are lower than those of adversarial examples generally.}
\label{klValuePixel}
\end{figure*}

We respectively select 50 random images from benign examples and each type of adversarial examples, and generate their transformed versions via pixel modification. Then, we compute the $D_{KL}$ values between selected examples and their transformed versions. As shown in Figure~\ref{klValuePixel}, the $D_{KL}$ values of benign and adversarial examples are plotted in color lines.

In Figure~\ref{klValuePixel}, the additive noise is realized by Gaussian noise, the maximum smoothing is a maximum filter with the $3 \times 3$ sliding window, the bit-depth of images is reduced from 8 bits to 4 bits, and Feature-filter, one of the variants of the frequency-domain transform, is reproduced by only reserving the $0.7 \times 0.7$ low-frequency coefficients as described in ~\cite{LiuH}. As shown in Figure~\ref{klValuePixel}, the $D_{KL}$ values between benign images and their transformed versions are usually located at a lower level. These approaches via pixel modification are very effective for C\&W attacks but are not always satisfactory against the other two attacks.

We test the performance of four groups of pixel modifications against three typical adversarial attacks. Table~\ref{table_2} lists the average $D_{KL}$ values between test images (benign images and adversarial images generated by C\&W, FGSM, and BIM attacks) and their transformed images. The average $D_{KL}$ values of adversarial examples are usually greater than those of benign images. The results infer that adversarial examples (especially those generated by C\&W attacks) are usually sensitive to pixel modifications, while benign images are immune to these operations. The detection rate under a certain threshold is also illustrated in Table~\ref{table_2}.

As listed in Table~\ref{table_2}, although C\&W attacks have been proven to produce more imperceptible adversarial examples, they are not robust enough to pixel modifications. Most of pixel modifications show an accuracy of more than 90\% in detecting adversarial examples generated by C\&W attacks. Adversarial examples generated by FGSM and BIM attack are more likely to escape the detector based on pixel modifications. The BIM attack, in particular, can escape detection with a very high probability. Therefore, we do not believe these results are good enough for use as a stand-alone detection. We also observe that the parameters of pixel modification have a significant effect on the detection rate. For the additive noise against C\&W attack, the accuracy is up to 100\% when the parameter is set to Speckle, while the accuracy is close to 0 when the parameter is set to Poisson. In general, the detectors based on pixel modification have an excellent ability to detect C\&W attacks, but are relatively poor in detecting the other two attacks with low perturbation levels.

\begin{table*}[htb]
\caption{Average $D_{KL}$ values and detection rate via pixel modification.\label{table_2}}
\centering
\begin{tabular}{|l||r||r||r||r||r||r||r||r||r||r|}
\hline
\multirow{2}{*}{Transformation} & \multirow{2}{*}{Parameter} & \multirow{2}{*}{Threshold} & \multicolumn{4}{c||}{Average $D_{KL}$} & \multicolumn{4}{c|}{Accuracy} \\
\cline{4-11}
&  &  & C\&W & FGSM & BIM & Benign & C\&W & FGSM & BIM & Benign \\

\hline

\multirow{4}*{Additive noise} & Gaussian & 0.273 & 3.123 & 0.392 & 1.416 & 0.391 & 98\% & 45\% & 44\% & \textbf{71\%} \\
                         %\cline{2-7}
                         & Poisson & 0.010 & 0 & 0.049 & 0.008 & 0.121 & 0\% & \textbf{71\%} & 5\% & 42\% \\
                         %\cline{2-7}
                         & Salt\&Pepper & 0.519 & 2.938 & 0.659 & 2.381 & 0.777 & 97\% & 42\% & \textbf{54\%} & 67\% \\
                         %\cline{2-7}
                         & Speckle & 0.071 & 2.641 & 0.127 & 0.093 & 0.184 & \textbf{100\%} & 48\% & 9\% & 68\% \\
\hline
\multirow{5}*{Smoothing} & Gaussian & 0.346 & 2.928 & 0.985 & 2.403 & 0.266 & \textbf{100\%} & 68\% & 56\% & 82\% \\
                                 & Maximum  & 0.710 & 2.846 & 1.811 & 5.450 & 0.460 & 96\% & 71\% & 88\% & 79\% \\
                                 & Median   & 0.319 & 2.961 & 1.133 & 2.012 & 0.159 & 99\% & \textbf{77\%} & 53\% & \textbf{88\%} \\
                                 & Uniform  & 0.375 & 2.906 & 1.019 & 2.311 & 0.248 & 99\% & 70\% & 53\% & 83\% \\
                                 & Minimum  & 0.940 & 2.849 & 1.914 & 5.643 & 0.533 & 91\% & 70\% & \textbf{90\%} & 83\% \\
\hline
\multirow{3}*{Bit-depth reduction} & 4-bit  & 0.020 & 2.602 & 0.032 & 0.014 & 0.062 & \textbf{100\%} & 41\% & \textbf{9\%} & 67\% \\
                         %\cline{2-7}
                         & 5-bit  & 0.004 & 1.179 & 0.006 & 0.011 & 0.012 & 99\% & 42\% & 8\% & 68\% \\
                         %\cline{2-7}
                         & 6-bit  & 0.001 & 0.358 & 0.002 & 0.001 & 0.002 & 98\% & \textbf{48\%} & 6\% & \textbf{71\%} \\
\hline
\multirow{4}*{Feature-filter} & 0.6 & 0.170 & 2.340 & 0.614 & 0.252 & 0.097 & 98\% & 69\% & \textbf{23\%} & 84\% \\
                              & 0.7 & 0.050 & 1.674 & 0.321 & 0.081 & 0.030 & \textbf{99\%} & 82\% & 13\% & 84\% \\
                              & 0.8 & 0.011 & 0.788 & 0.118 & 0.022 & 0.005 & \textbf{99\%} & \textbf{87\%} & 13\% & 87\% \\
                              & 0.9 & 0.003 & 0.221 & 0.025 & 0.006 & 0.001 & 98\% & 84\% & 13\% & \textbf{92\%} \\
\hline
\end{tabular}

\end{table*}

%-------------------------------------------------------------------------
\subsection{Topological Transformation Tests}
\begin{figure*}[htp]
\centering
\subfloat[Translation]{
    \includegraphics[width=0.64\columnwidth]{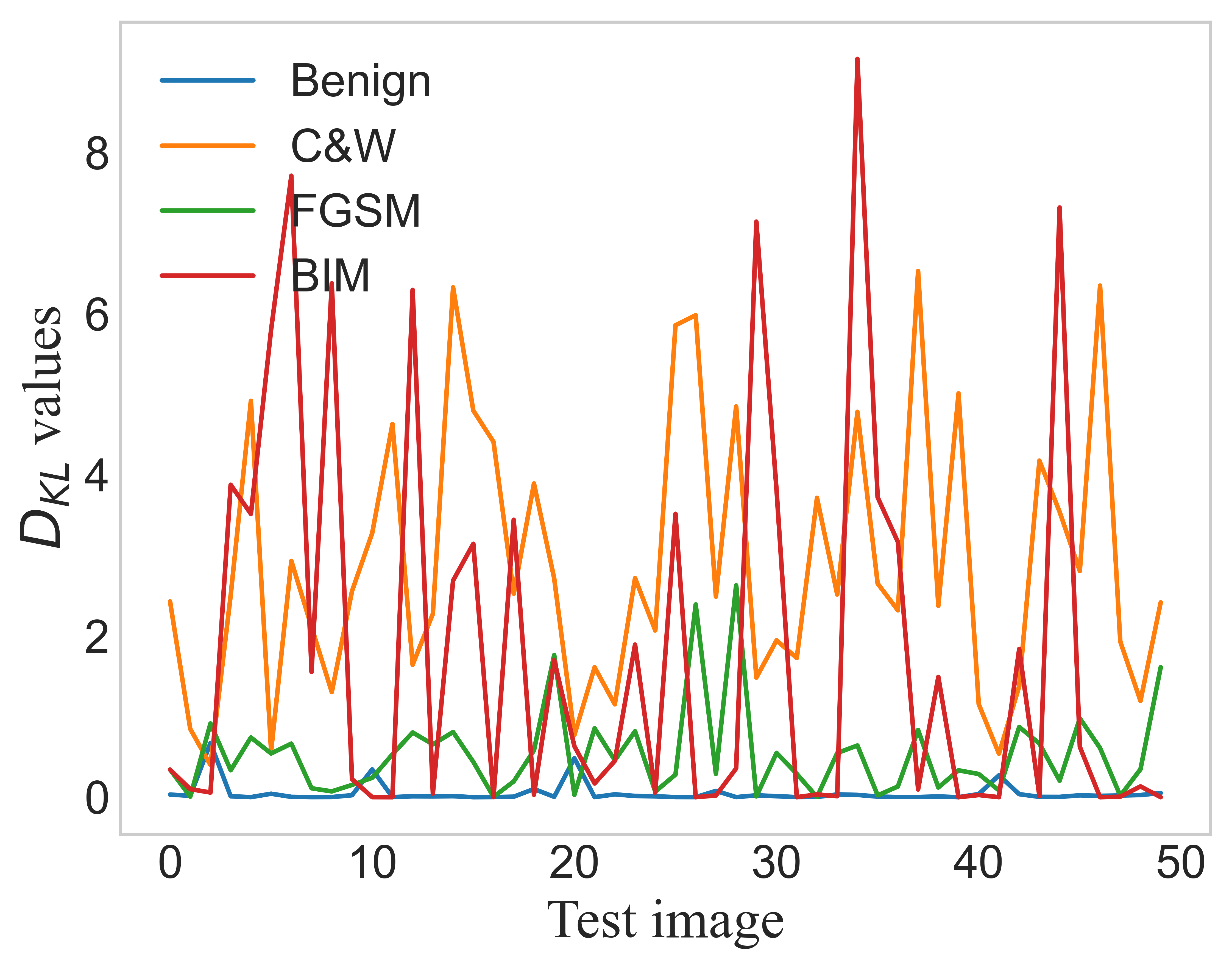}
}   
\subfloat[Flip]{
    \includegraphics[width=0.64\columnwidth]{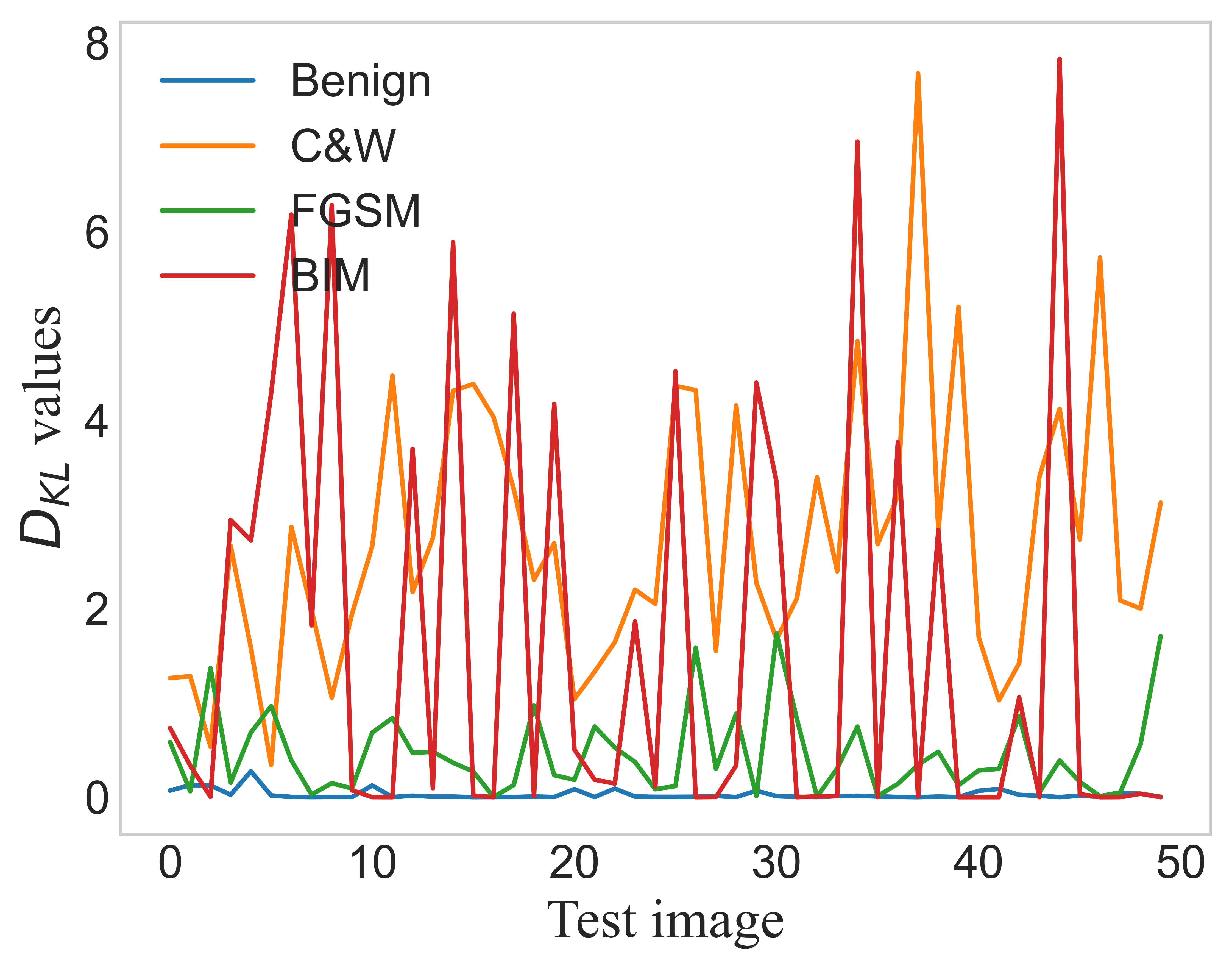}
} \\
\subfloat[Rotation]{
    \includegraphics[width=0.64\columnwidth]{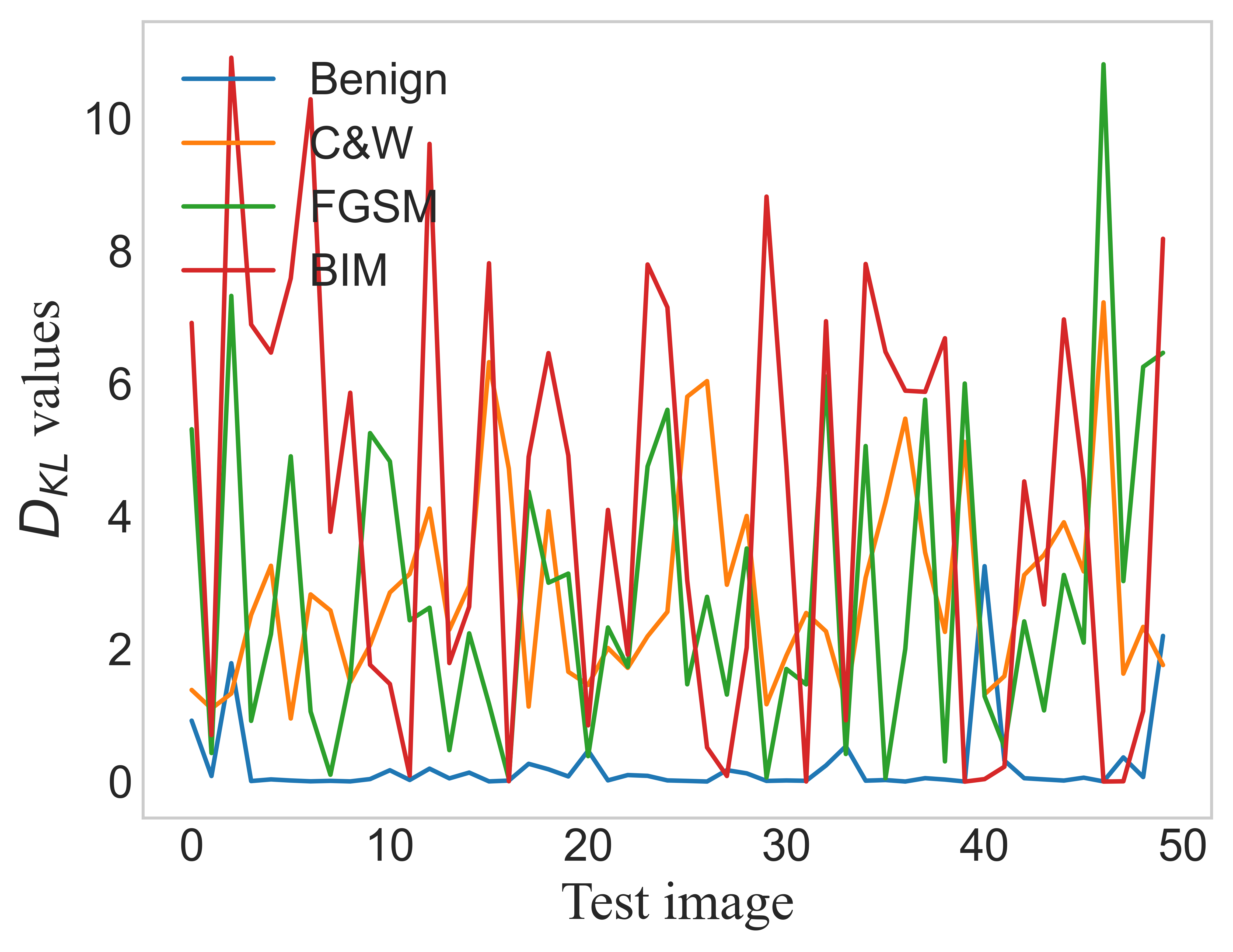}
}
\subfloat[Shear]{
    \includegraphics[width=0.64\columnwidth]{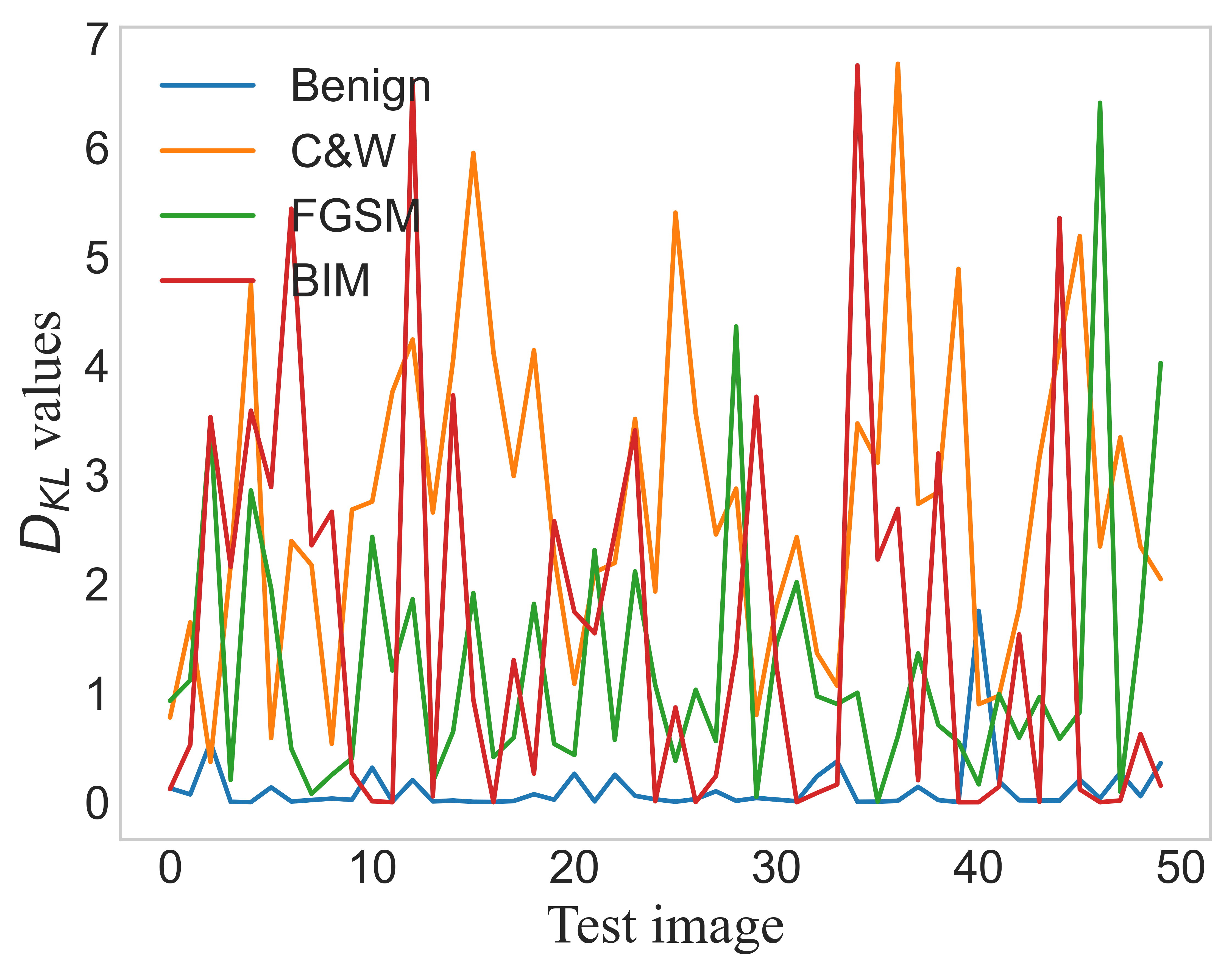}
}
\subfloat[Scale]{
    \includegraphics[width=0.64\columnwidth]{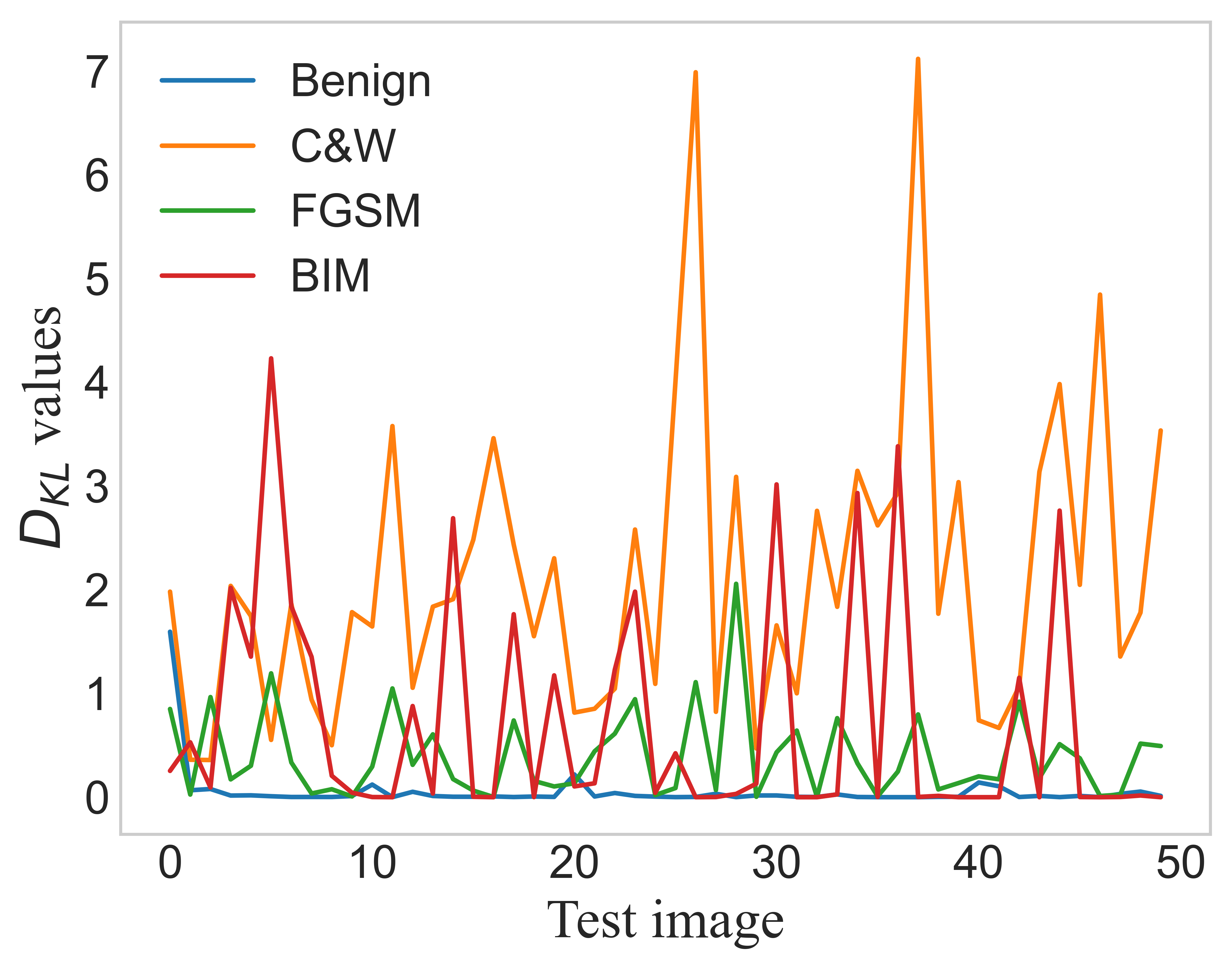}
}
\caption{Distributions of the $D_{KL}$ values via topological transformation. The $D_{KL}$ values of benign images are lower than those of adversarial examples generally.}
\label{klValueTopol}
\end{figure*}

As shown in Figure~\ref{klValueTopol}, we also select 50 random images, and compute the $D_{KL}$ values between test images (benign and adversarial images) and their transformed versions via topological transformation. The $D_{KL}$ values of benign images are plotted in blue lines, and those of adversarial images are plotted in other colors. 

In Figure~\ref{klValueTopol}, the translation parameters are set to $ [\alpha, \beta]^T = [5, 5]^T $, the warping matrix $T = \left[ \begin{matrix} -1 & 0 \\ 0 & 1 \end{matrix} \right] $ indicates the horizontal flipping of the test image, the rotation angle $\theta$ is set to $20^{\circ}$, and the shear parameter $ a = 0.3 $. The scale parameter $ s = 0.8 $ indicates that the test images are shrunk by 0.8 times and then enlarged by 1.25 times to retain the same scale. This set of figures shows $D_{KL}$ distributions of benign images are visibly lower than those of adversarial examples. The results indicate that topological transformations could help to detect adversarial examples.

Similarly, we test the effect of topological transformations on adversarial detection with a given set of parameters. Table~\ref{table_3} lists the average $D_{KL}$ values of the test images and the detection rate under a certain threshold. The average $D_{KL}$ values of adversarial examples are always greater than those of benign examples. This results suggest that adversarial examples are more sensitive to topological transformation than benign examples. Among these adversarial attacks, C\&W attacks is most likely to be detected by the detector base on topological transformations, followed by the FGSM attack. The BIM attack can still escape detection with a high probability. In addition to the threshold, the parameters of topological transformation can affect the accuracy of adversarial detectors. For example, horizontal and vertical flips show significant difference in detection rate. We also observe that the image scale achieves a high accuracy of up to 100\% against C\&W attacks, but poor performance against the BIM attack. This results indicate that adversarial detection based on an individual topological transformation is still not reliable. No individual kind of topological transformation has generic detection ability. In particular, most adversarial examples based on the BIM attack can evade these detectors.

\begin{table*}[htb]
\caption{Average $D_{KL}$ values and detection rate via topological transformation.\label{table_3}}
\centering
\begin{tabular}{|l||r||r||r||r||r||r||r||r||r||r|}
\hline
\multirow{2}{*}{Transformation} & \multirow{2}{*}{Parameter} & \multirow{2}{*}{Threshold} & \multicolumn{4}{c||}{Average $D_{KL}$} & \multicolumn{4}{c|}{Accuracy} \\
\cline{4-11}
&  &  & C\&W & FGSM & BIM & Benign & C\&W & FGSM & BIM & Benign \\
\hline
\multirow{5}*{Translation} & $ [1, 1]^T $ & 0.083 & 2.618 & 0.461 & 0.455 & 0.030 & \textbf{100\%} & \textbf{78\%} & 32\% & 91\% \\
                         %\cline{2-7}
                         & $ [3, 3]^T $ & 0.228 & 2.982 & 0.607 & 0.981 & 0.056 & 99\% & 67\% & \textbf{44\%} & \textbf{93\%} \\
                         %\cline{2-7}
                         & $ [5, 5]^T $ & 0.227 & 3.039 & 0.678 & 1.202 & 0.067 & 98\% & 69\% & \textbf{44\%} & 92\% \\
                         %\cline{2-7}
                         & $ [7, 7]^T $ & 0.193 & 3.004 & 0.685 & 0.994 & 0.069 & \textbf{100\%} & 70\% & 42\% & 89\% \\
                         %\cline{2-7}
                         & $ [9, 9]^T $ & 0.224 & 3.034 & 0.713 & 1.062 & 0.078 & 98\% & 67\% & 41\% & 88\% \\
\hline
\multirow{3}*{Flip} & Vertical & 1.095 & 2.646 & 2.402 & 2.622 & 1.352 & 93\% & 77\% & 65\% & 57\% \\
                    & Horizontal & 0.130 & 3.003 & 0.601 & 0.820 & 0.056 & \textbf{99\%} & \textbf{80\%} & 41\% & \textbf{88\%} \\
                    & Both  & 0.979 & 2.608 & 2.415 & 2.444 & 1.343 & 93\% & 79\% & \textbf{66}\% & 58\% \\
\hline
\multirow{6}*{Rotation} & $-30^{\circ}$  & 1.021 & 2.886 & 3.107 & 4.159 & 0.781 & 90\% & 83\% & 78\% & 75\% \\
                         %\cline{2-7}
                         & $-20^{\circ}$  & 0.861 & 2.952 & 2.658 & 3.331 & 0.448 & 94\% & \textbf{78\%} & 72\% & 85\% \\
                         %\cline{2-7}
                         & $-10^{\circ}$  & 0.586 & 3.029 & 2.090 & 1.823 & 0.205 & \textbf{96\%} & 74\% & 58\% & \textbf{90\%} \\
                         & $10^{\circ}$  & 0.420 & 3.043 & 2.070 & 1.803 & 0.204 & \textbf{96\%} & \textbf{78\%} & 52\% & 87\% \\
                         %\cline{2-7}
                         & $20^{\circ}$  & 1.049 & 2.978 & 2.684 & 3.339 & 0.445 & 88\% & \textbf{74\%} & 65\% & 87\% \\
                         %\cline{2-7}
                         & $30^{\circ}$  & 1.003 & 2.951 & 3.178 & 4.269 & 0.740 & 93\% & 83\% & 79\% & 73\% \\
\hline
\multirow{5}*{Shear} & 0.1 & 0.165 & 2.639 & 0.672 & 0.339 & 0.085 & \textbf{98\%} & 76\% & 26\% & \textbf{85\%} \\
                     & 0.2 & 0.311 & 2.719 & 0.969 & 0.644 & 0.161 & \textbf{98\%} & 73\% & 39\% & \textbf{85\%} \\
                     & 0.3 & 0.478 & 2.784 & 1.381 & 1.302 & 0.303 & 98\% & 78\% & 52\% & 83\% \\
                     & 0.4 & 0.660 & 2.785 & 1.865 & 2.225 & 0.502 & 96\% & \textbf{80\%} & 63\% & 80\% \\
                     & 0.5 & 0.884 & 2.760 & 2.401 & 3.253 & 0.793 & 93\% & \textbf{80\%} & \textbf{71\%} & 72\% \\
\hline
\multirow{4}*{Scale} & 0.8 & 0.127 & 2.404 & 0.531 & 0.260 & 0.063 & \textbf{100\%} & 73\% & \textbf{22\%} & \textbf{89\%} \\
                     & 0.9 & 0.109 & 2.144 & 0.391 & 0.142 & 0.048 & \textbf{100\%} & 70\% & 17\% & 88\% \\
                     & 1.1 & 0.070 & 1.754 & 0.290 & 0.082 & 0.032 & \textbf{100\%} & 73\% & 13\% & 87\% \\
                     & 1.2 & 0.050 & 1.622 & 0.277 & 0.070 & 0.027 & \textbf{100\%} & \textbf{75\%} & 8\% & \textbf{89\%} \\
\hline
\end{tabular}
\end{table*}

%------------------------------------------------------------------------
\subsection{Summary}
In this section, we classify the 9 image transformations in adversarial detection into two dimensions, and test their detection performances based on the baseline architecture. A total of 1,440 samples are used to evaluation in the detection experiments, half of them are benign and half are adversarial. These adversarial examples against the Inception V3 model are generated by C\&W, FGSM, and BIM attacks on the ImageNet dataset.

By analyzing the results of the pixel modification and topological transformation, we find that individual transformations are helpful for adversarial detection, but not reliable enough. These transformations are generally effective in detecting C\&W attacks, but mediocre against FGSM, and BIM attacks. There are also a high false positive rate (FPR) for benign examples. Compared with pixel modification, topological transformation is more effective in detecting adversarial examples in most cases, especially those generated by the BIM attack.

Since the baseline detection architecture modifies inputs rather than the target DNN model, it is inexpensive and easy to combine with other defenses. However, the detection performance of this architecture heavily relies on the threshold setting. The good detector built on the baseline architecture could take a lot of time to find an appropriate threshold. In addition, the parameters of image transformation can also affect the performance of adversarial detector.

Because of the limited detection ability of individual transformations, and the difficulty of setting an appropriate threshold for the baseline architecture, we seek to combine multiple image transformations based on a new detection architecture, aiming to improve the performance of the adversarial detector.

%------------------------------------------------------------------------
\section{Joint Detection}
\subsection{Our Approach}
Adversarial detector based on an individual image transformation cannot achieve satisfactory accuracy. How to improve the detection ability by combining the $D_{KL}$ values of multiple image transformations? To answer this question, we design a DNN-based detector referred as to \emph{AdvJudge}.

\emph{AdvJudge} is constructed by a simple DNN model. The implementation of this detector includes 3 steps: data preprocessing, constructing and training a DNN, and testing.

\begin{figure}[htb]
\centering
\includegraphics[width=0.96\columnwidth]{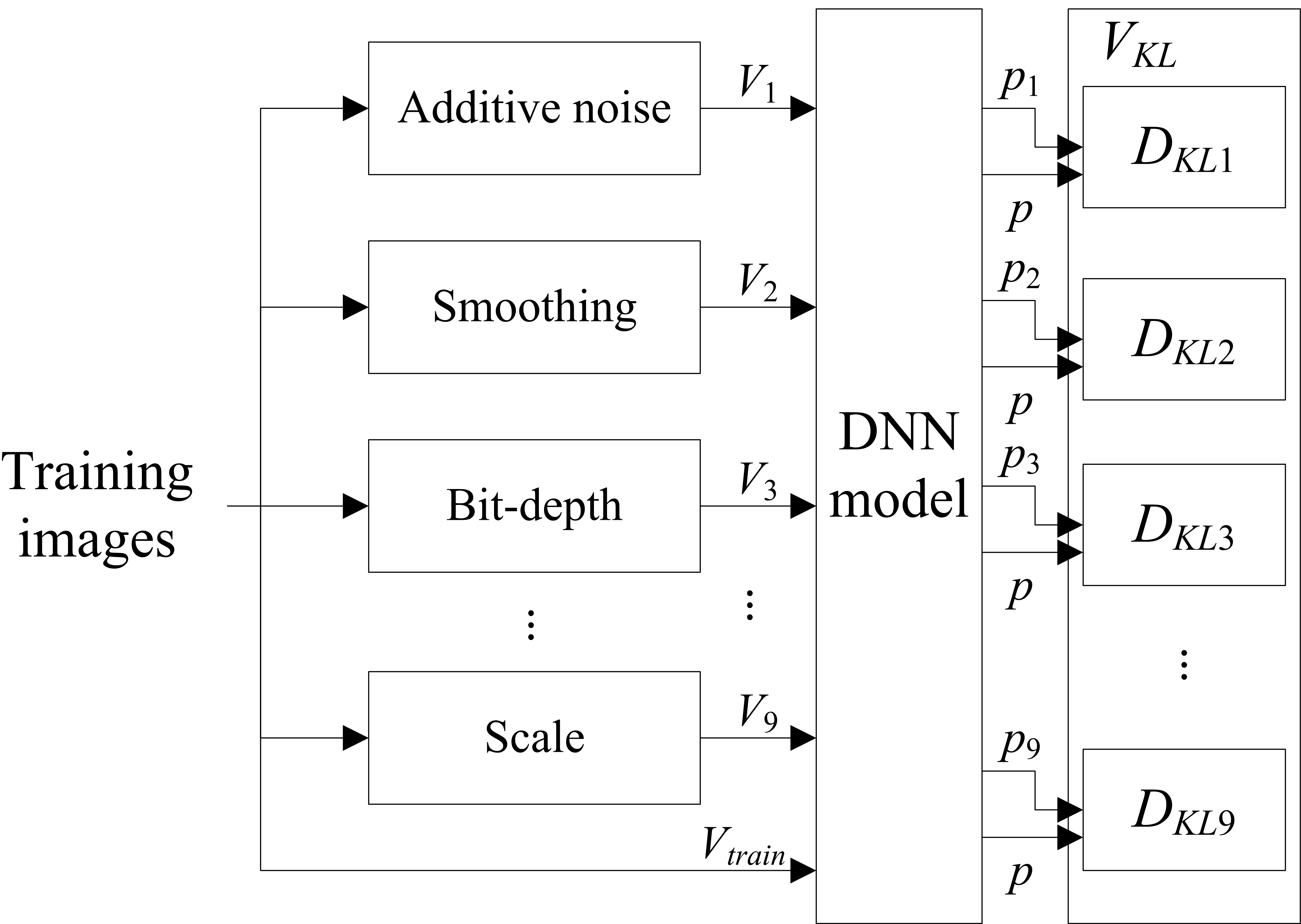}
\caption{Data preprocessing. Nine types of image transformations are preformed to form a feature vector with 9 $D_{KL}$ values.}
\label{preprocessing}
\end{figure}

(1) Data preprocessing. We combine 9 image transformations to obtain transformed versions, and calculates $D_{KL}$ values between training images and their transformed versions. The framework of data preprocessing is plotted in Figure~\ref{preprocessing}. For each training image, a feature vector $V_{KL}$ consists of 9 $D_{KL}$ values. If the feature vector comes from an adversarial image, it is labeled 1; otherwise, it is labeled 0.

(2) Constructing and training a DNN. \emph{AdvJudge} is a simple neural network that consists of fully connected layers. Details of this neural network are listed in Tab~\ref{detector_DNN}. This neural network is trained by the above feature vectors with labels.

(3) Testing. The test image is preprocessed as shown in Figure~\ref{preprocessing} to generate its feature vector. \emph{AdvJudge} accepts the feature vector and gives a positive predictive value $\rho$. If $\rho > 0.5$, the test image is considered adversarial; otherwise, it is considered benign. 

\begin{table}[htb]
\caption{Our detector architecture.\label{detector_DNN}}
\centering
\begin{tabular}{|l||r||r||r|}
\hline
Layer & Type & Activation & Output shape \\
\hline
Input  &  &  &  9 \\
1  & FC & Relu &  64 \\
2  & FC & Relu &  64 \\
3  & FC & Relu &  64 \\
4  & FC & Relu &  64 \\
5  & FC & Sigmoid &  1 \\
\hline
\end{tabular}

\end{table}

\subsection{Results}
We utilize the same experimental setup in subsection IV-A to train \emph{AdvJudge}, and compare its performance with individual transformations. Table~\ref{scores} shows the detection ability for several detectors built upon individual image transformations and our joint detector under the threshold of 0.5. We employ the precision and recall~\cite{LiangB} to evaluate the ability of \emph{AdvJudge}. The precision is the ratio $\frac{tp}{tp + fp}$ where $ tp $ is the number of true positives and $ fp $ the number of false positives. The precision is intuitively the ability of the classifier not to label as positive a sample that is negative. The recall is the ratio $ \frac{tp}{tp + fn} $ where $ tp $ is the number of true positives and $ fn $ the number of false negatives. The recall is intuitively the ability of the classifier to find all the positive samples. The $F1$ score~\cite{LiangB} is a harmonic mean of the precision and recall, where an $F1$ score reaches its best value at 1 and worst score at 0. The $F1$ score is widely used to measure the success of a binary classifier, which can be formulated as follows:

\begin{equation}
F1 = \frac{2 \times precision \times recall}{precision + recall}
\label{f1}
\end{equation}

As listed in Table~\ref{scores}, among individual image transformations, smoothing with the maximum filter and rotation with the angle $\theta = -10^{\circ}$ achieve the best performance up to a 0.83 $F1$ score, while bit-depth reduction to 6 achieves the worst performance with with a 0.56 $F1$ score. We believe that even the best performance of individual transformations is not sufficient to reliably detect adversarial examples. \emph{AdvJudge} reports 92\% precision and 94\% recall, which are significant higher than those based on other individual transformations. The $F1$ score of our detector is 0.92 and achieves a 9\% improvement compared to the best result among individual transformations.

\begin{table}[htb]
\caption{Performance of detectors.\label{scores}}
\centering
\begin{tabular}{|l||r||r||r||r|}
\hline
Transformation & Parameter & Precision & Recall & $F1$ score \\
\hline
Additive noise & Gaussian & 64\% & 66\% & 0.65  \\
Smoothing  & Maximum & 85\% & 83\% & 0.83  \\
Bit-depth reduction & 6 & 50\% & 64\% & 0.56   \\
Feature-filter  & 0.9 & 65\% & 89\% & 0.75   \\
Translation  & 1 & 70\% & 91\% & 0.79   \\
Flip  & Horizontal & 73\% & 86\% & 0.79   \\
Rotation  & $-10^{\circ}$ & 81\% & 84\% & 0.83   \\
Shear & 0.3 & 76\% & 81\% & 0.78   \\
Scale & 1.1 & 65\% & 85\% & 0.74  \\
\emph{AdvJudge} & All & \textbf{92\%} & \textbf{94\%} & \textbf{0.92}   \\
\hline
\end{tabular}
\end{table}

\begin{figure}[htb]
\centering
\includegraphics[width=0.85\columnwidth]{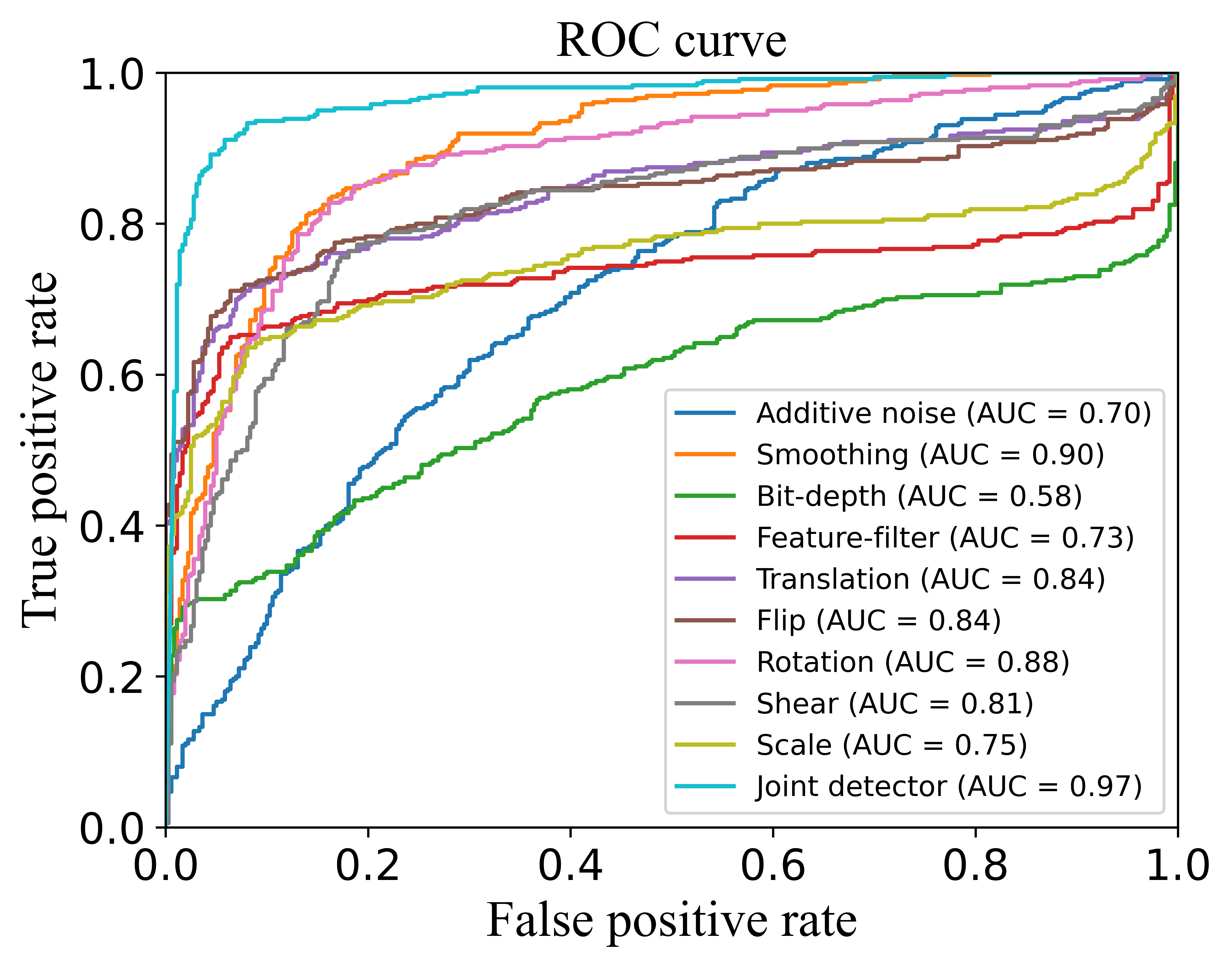}
\caption{ROC curves for our \emph{AdvJudge} and other adversarial detectors based on image transformations.}
\label{roc}
\end{figure}

A receiver operating characteristic (ROC) curve~\cite{Kherchouche} is a graphical plot that illustrates the detection ability of a binary classifier as its threshold is varied. The ROC curve is created by plotting TPR against FPR at various threshold settings. Figure~\ref{roc} shows ROC curves for detectors based on each image transformation and the joint transformation. The ROC curve of the joint detector is always higher than that of the others, indicating that \emph{AdvJudge}'s performance is better than those based on individual transformations under all thresholds. To measure the performance of detectors, we calculate the area under the curve (AUC), where an AUC value varies between 0 and 1, with an uninformative detector yielding 0.5. We list the AUC value of all detectors in Figure~\ref{roc}. In terms of the performance of detectors based on individual transformations, smoothing with the maximum filter achieves the highest AUC value of up to 0.90, while bit-depth reduction to 6 only achieves an AUC value of 0.58. Different image transformations have significant differences in the performance of the adversarial detector. Therefore, it is important to select an appropriate image transformation against adversarial examples. As shown in Figure~\ref{roc}, \emph{AdvJudge} reports an AUC value of up to 0.97, which is significantly higher than the results among individual transformations. Therefore, we suggest that \emph{AdvJudge} is able to effectively detect adversarial examples from a wide range of attacks.

\subsection{Interpretability}
Beyond considering the accuracy of our \emph{AdvJudge}, it is important to understand how this detector is working and it's decision. To answer this question, we utilize the Captum interpretability library to understand which image transformations are more influential and how our \emph{AdvJudge} reaches its prediction. Captum is a DNN interpretability and understanding library for PyTorch, which contains general implementations of several state-of-the-art interpretability methods~\cite{Sundararajana, Sundararajanb, shrikumar, Tomsett, Qudrat}. As a main interpretability method, the integrated gradient~\cite{Sundararajana, Sundararajanb} provide an easy way to understand which image transformations are contributing more to the \emph{AdvJudge}'s decision. The integrated gradient IGs along the $i'th$ dimension for an input $x$ to baseline $x'$ is defined as follows:
\begin{equation}
\text {IGs}_{i}(x)::=\left(x_{i}-x_{i}'\right) \times \int_{\alpha=0}^{1} \frac{\partial f\left(x^{\prime}+\alpha \times\left(x-x^{\prime}\right)\right)}{\partial x_{i}} d \alpha
\label{IntegratedGrads}
\end{equation}
where $\frac{\partial f(x)}{\partial x_{i}}$ is the gradient of a DNN model $f(x)$ along the $i'th$ dimension.

\begin{figure}[htb]
\centering
\includegraphics[width=0.9\columnwidth]{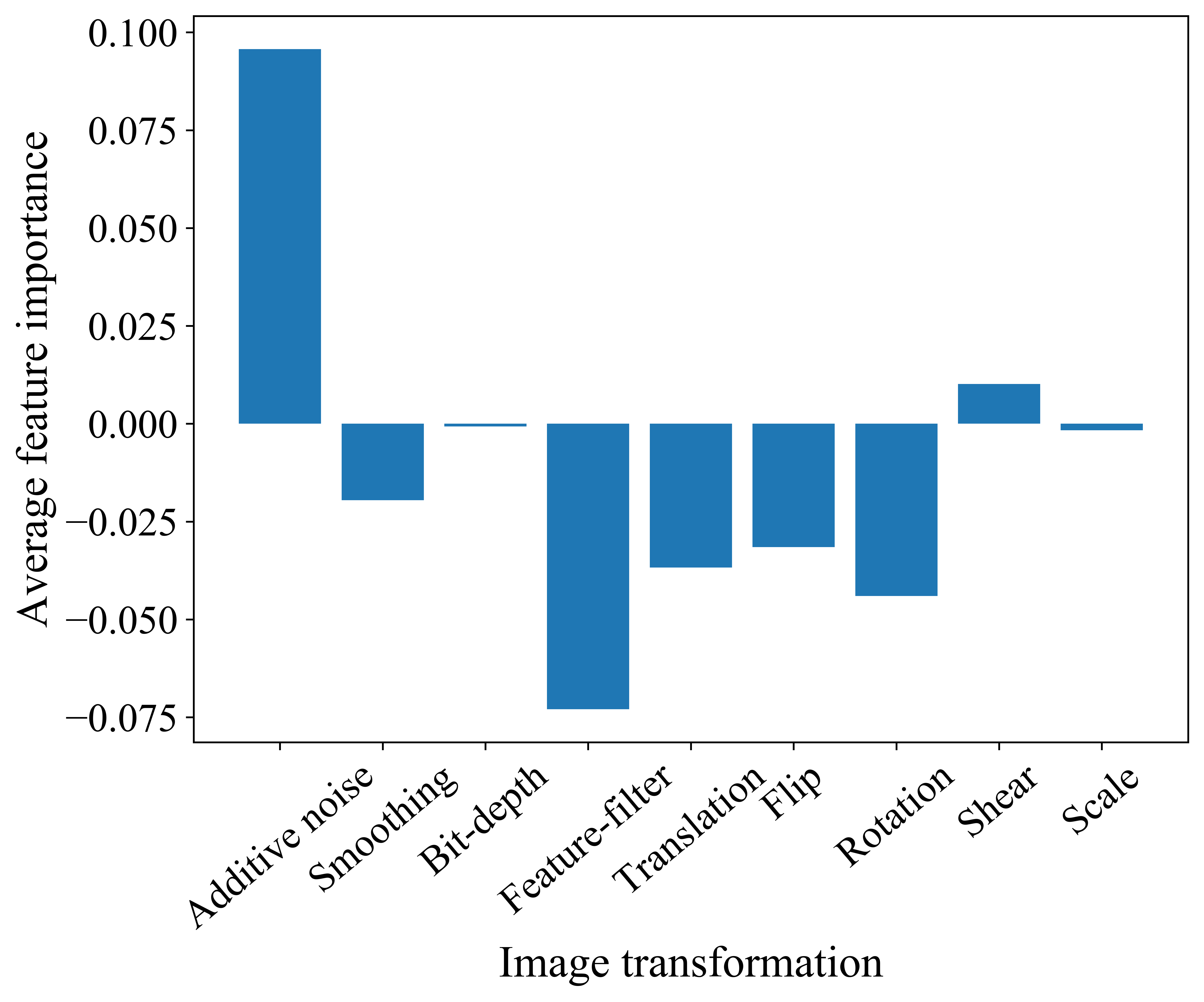}
\caption{Average feature importance for identifying benign images. The strongest feature appears to be additive noise.}
\label{feature_importances_Ben}
\end{figure}

\begin{figure}[htb]
\centering
\includegraphics[width=0.9\columnwidth]{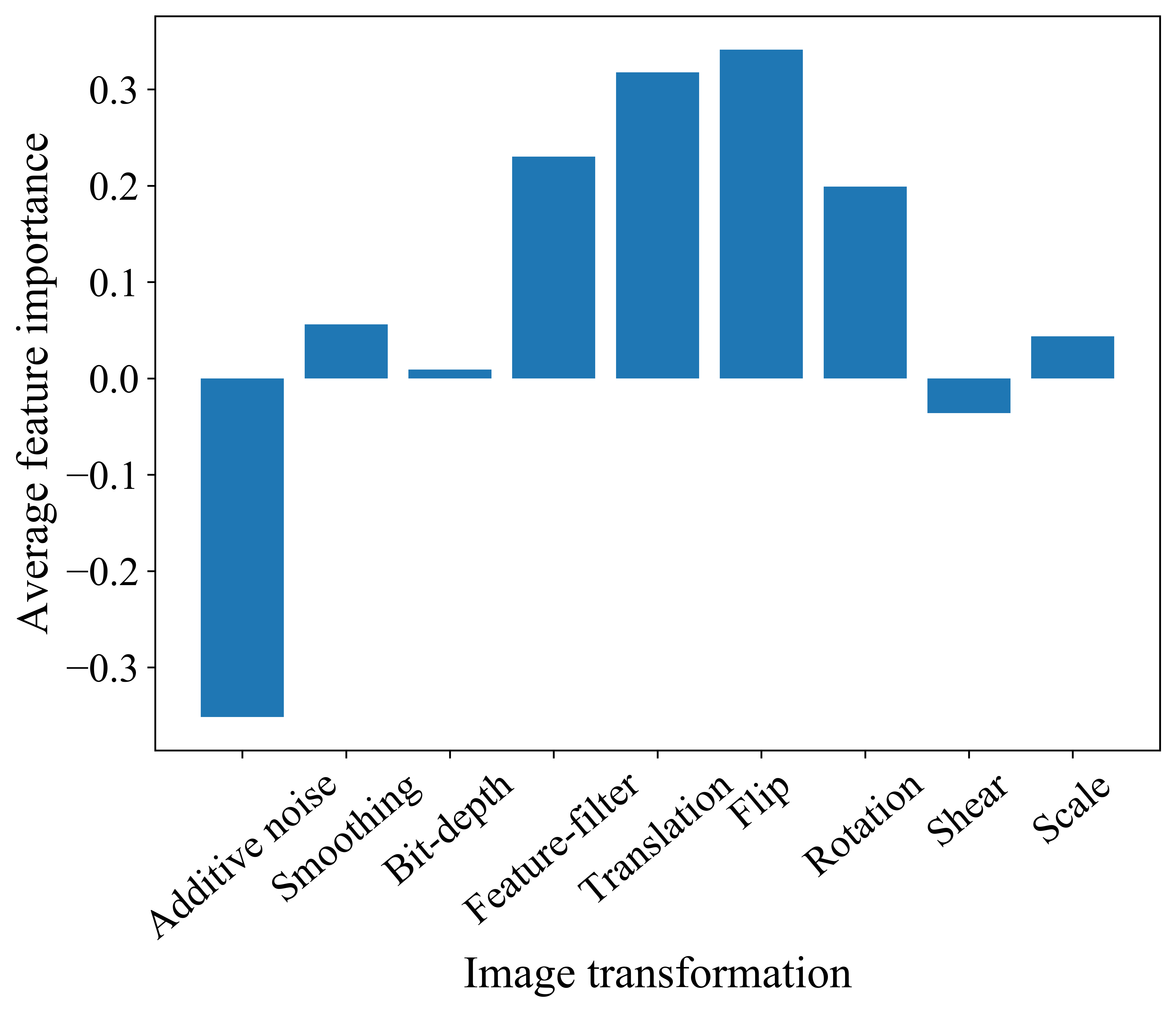}
\caption{Average feature importance for identifying adversarial images. Bit-depth appear to be less important feature.}
\label{feature_importances_Adv}
\end{figure}

In order to show contributions of each image transformation to adversarial detection, we average these integrated gradients across all the test images and visualize them for each image transformation. Figure~\ref{feature_importances_Ben} and Figure~\ref{feature_importances_Adv} show the average feature importance for identifying benign and adversarial images, respectively. We find that the $D_{KL}$ values of additive noise and shear are positively correlated with benign image prediction, while those of the other transformations are positively correlated with adversarial detection. In terms of relative importance, we observe that the influence of addition noise is about 9.5 times as great as shear when identifying benign images; the influence of flip is the greatest and the influence of bit-depth is minimum when detecting adversarial examples. The combination of these image transformations forms a generic detection ability for our detector to improve the precision and recall.

\subsection{Case Study}
What is the primary cause for each false positive or false negative? To answer this question, we conduct a case study for one false positive and each type of false negative. The false-positive case is a benign example but judged by \emph{AdvJudge} to be adversarial. The false-negative cases are adversarial examples generated by C\&W, FGSM, and BIM attacks, but they are judged to be benign. Table~\ref{caseStudy1} lists the $D_{KL}$ values of each image transformation and predictions of \emph{AdvJudge} in the four cases. In the false-negative cases, \emph{AdvJudge} gives lower predictive values than the threshold 0.5, thus they are wrongly judged to be benign. In the false-positive case, \emph{AdvJudge} gives a predictive value close to 1, thus they are wrongly judged to be adversarial.

\begin{table}[htb]
\caption{$D_{KL}$ values of each image transformation and predictions of \emph{AdvJudge} in the four cases.\label{caseStudy1}}
\centering
\begin{tabular}{|l||r||r||r||r|}
\hline
\multirow{2}{*}{Transformation} & \multicolumn{3}{c||}{False-negative} & \multirow{2}{*}{False-positive} \\
\cline{2-4}
 & C\&W & FGSM & BIM & \\
\hline
Additive noise & 2.938 & 1.378 & 0.763 & 0.411  \\
Smoothing  & 1.425 & 0.254 & 0.930 & 0.405  \\
Bit-depth reduction & 0.021 & 0.004 & 0.0 & 0.043   \\
Feature-filter  & 1.425 & 0.254 & 0.930 & 0.405   \\
Translation  & 0.910 & 0.439 & 0.061 & 0.804   \\
Flip  & 0.851 & 0.107 & 0.066 & 0.830   \\
Rotation  & 0.640 & 0.555 & 0.416 & 2.232   \\
Shear & 1.626 & 0.634 & 0.865 & 0.735   \\
Scale & 1.584 & 0.196 & 0.022 & 0.745  \\
Predictive value & 0.154 & 0.062 & 0.075 & 0.998 \\
\hline
\end{tabular}
\end{table}

We employ the integrated gradient tool to analyze the contributions of each $D_{KL}$ value to the \emph{AdvJudge} detector. Table~\ref{caseStudy2} lists the results of each image transformation to adversarial detection of the false positive and false negatives. In false-negative cases, the $D_{KL}$ value of additive noise has a significant negative correlation with adversarial detection, which results in a fail detection. In the false-positive case, the $D_{KL}$ values of translation and flip have a significant negative correlation with benign example prediction, while the $D_{KL}$ value of additive noise gives a relatively small positive contribution. These transformations, especially flip, translation and rotation, together lead to false alarms of the \emph{AdvJudge} detector.

\begin{table}[htb]
\caption{Contributions of each image transformation to adversarial detection of the false positive and false negatives.\label{caseStudy2}}
\centering
\begin{tabular}{|l||r||r||r||r|}
\hline
\multirow{2}{*}{Transformation} & \multicolumn{3}{c||}{False-negative} & \multirow{2}{*}{False-positive} \\
\cline{2-4}
 & C\&W & FGSM & BIM & \\
\hline
Additive noise & \textbf{-4.624} & \textbf{-0.193} & \textbf{-0.121} & 0.147  \\
Smoothing  & -0.064 & 0.006 & 0.025 & -0.030  \\
Bit-depth reduction & 0.023 & 0.001 & 0.0 & -0.009   \\
Feature-filter  & 0.478 & 0.026 & 0.106 & -0.040   \\
Translation  & 1.840 & 0.131 & 0.021 & -0.365   \\
Flip  & 1.491 & 0.020 & 0.014 & \textbf{-0.491}   \\
Rotation  & 0.548 & 0.109 & 0.093 & -0.110   \\
Shear & -0.384 & -0.066 & -0.102 & 0.001   \\
Scale & 0.804 & -0.012 & -0.002 & -0.046  \\
\hline
\end{tabular}
\end{table}

\section{Conclusion}
Image transformations help to detect adversarial examples. Such approaches are simple and inexpensive, and do not modify the architectures or parameters of the neural network. Therefore, a detector based on image transformations can be employed to detect adversarial examples at runtime without accuracy loss of the neural network. In this paper, we explore and test the effect of image transformation on adversarial examples generated by state-of-the-art adversarial attacks. The results demonstrate that an individual transformation has difficulty detecting adversarial examples in a robust way. This motivates us to develop a joint approach to improve detectability. We combine 9 image transformations to obtain their scores, and construct a set of feature vectors. A simple DNN learns these feature vectors to become a good judge. Experimental results show that our \emph{AdvJudge} achieves a 0.92 $F1$ score and an AUC value of up to 0.97, which significantly outperforms all detectors based on an individual transformation. Beyond considering the accuracy of \emph{AdvJudge}, we leverage an explainable AI tool to analyze the feature importance of each image transformation for adversarial detection. There are different contributions of each image transformation to benign and adversarial identification. The combination of them significantly improves the generic detection ability for adversarial detection.

\bibliographystyle{IEEEtran}
\bibliography{tifs22}

\begin{thebibliography}{10}
\providecommand{\url}[1]{#1}
\csname url@samestyle\endcsname
\providecommand{\newblock}{\relax}
\providecommand{\bibinfo}[2]{#2}
\providecommand{\BIBentrySTDinterwordspacing}{\spaceskip=0pt\relax}
\providecommand{\BIBentryALTinterwordstretchfactor}{4}
\providecommand{\BIBentryALTinterwordspacing}{\spaceskip=\fontdimen2\font plus
\BIBentryALTinterwordstretchfactor\fontdimen3\font minus
  \fontdimen4\font\relax}
\providecommand{\BIBforeignlanguage}[2]{{%
\expandafter\ifx\csname l@#1\endcsname\relax
\typeout{** WARNING: IEEEtran.bst: No hyphenation pattern has been}%
\typeout{** loaded for the language `#1'. Using the pattern for}%
\typeout{** the default language instead.}%
\else
\language=\csname l@#1\endcsname
\fi
#2}}
\providecommand{\BIBdecl}{\relax}
\BIBdecl

\bibitem{Carlini}
\BIBentryALTinterwordspacing
N.~Carlini and D.~A. Wagner, ``Towards evaluating the robustness of neural
  networks,'' in \emph{2017 {IEEE} Symposium on Security and Privacy, {SP}
  2017, San Jose, CA, USA, May 22-26, 2017}, 2017, pp. 39--57. [Online].
  Available: \url{https://doi.org/10.1109/SP.2017.49}
\BIBentrySTDinterwordspacing

\bibitem{Goodfellow}
\BIBentryALTinterwordspacing
I.~J. Goodfellow, J.~Shlens, and C.~Szegedy, ``Explaining and harnessing
  adversarial examples,'' in \emph{3rd International Conference on Learning
  Representations, {ICLR} 2015, San Diego, CA, USA, May 7-9, 2015, Conference
  Track Proceedings}, 2015. [Online]. Available:
  \url{http://arxiv.org/abs/1412.6572}
\BIBentrySTDinterwordspacing

\bibitem{Kurakin}
\BIBentryALTinterwordspacing
A.~Kurakin, I.~J. Goodfellow, and S.~Bengio, ``Adversarial examples in the
  physical world,'' in \emph{5th International Conference on Learning
  Representations, {ICLR} 2017, Toulon, France, April 24-26, 2017, Workshop
  Track Proceedings}, 2017. [Online]. Available:
  \url{https://openreview.net/forum?id=HJGU3Rodl}
\BIBentrySTDinterwordspacing

\bibitem{DuanY}
\BIBentryALTinterwordspacing
Y.~Duan, J.~Zou, X.~Zhou, W.~Zhang, J.~Zhang, and Z.~Pan, ``Enhancing
  transferability of adversarial examples via rotation-invariant attacks,''
  \emph{{IET} Comput. Vis.}, vol.~16, no.~1, pp. 1--11, 2022. [Online].
  Available: \url{https://doi.org/10.1049/cvi2.12054}
\BIBentrySTDinterwordspacing

\bibitem{ChenS}
\BIBentryALTinterwordspacing
S.~Chen, Z.~He, C.~Sun, J.~Yang, and X.~Huang, ``Universal adversarial attack
  on attention and the resulting dataset damagenet,'' \emph{{IEEE} Trans.
  Pattern Anal. Mach. Intell.}, vol.~44, no.~4, pp. 2188--2197, 2022. [Online].
  Available: \url{https://doi.org/10.1109/TPAMI.2020.3033291}
\BIBentrySTDinterwordspacing

\bibitem{HeK}
\BIBentryALTinterwordspacing
K.~He, X.~Zhang, S.~Ren, and J.~Sun, ``Deep residual learning for image
  recognition,'' in \emph{2016 {IEEE} Conference on Computer Vision and Pattern
  Recognition, {CVPR} 2016, Las Vegas, NV, USA, June 27-30, 2016}, 2016, pp.
  770--778. [Online]. Available: \url{https://doi.org/10.1109/CVPR.2016.90}
\BIBentrySTDinterwordspacing

\bibitem{HuangG}
\BIBentryALTinterwordspacing
G.~Huang, Z.~Liu, and K.~Q. Weinberger, ``Densely connected convolutional
  networks,'' \emph{CoRR}, vol. abs/1608.06993, 2016. [Online]. Available:
  \url{http://arxiv.org/abs/1608.06993}
\BIBentrySTDinterwordspacing

\bibitem{Karen}
\BIBentryALTinterwordspacing
K.~Simonyan and A.~Zisserman, ``Very deep convolutional networks for
  large-scale image recognition,'' in \emph{3rd International Conference on
  Learning Representations, {ICLR} 2015, San Diego, CA, USA, May 7-9, 2015,
  Conference Track Proceedings}, 2015. [Online]. Available:
  \url{http://arxiv.org/abs/1409.1556}
\BIBentrySTDinterwordspacing

\bibitem{HuJ}
\BIBentryALTinterwordspacing
J.~Hu, L.~Shen, and G.~Sun, ``Squeeze-and-excitation networks,'' \emph{CoRR},
  vol. abs/1709.01507, 2017. [Online]. Available:
  \url{http://arxiv.org/abs/1709.01507}
\BIBentrySTDinterwordspacing

\bibitem{Szegedy}
\BIBentryALTinterwordspacing
C.~Szegedy, W.~Zaremba, I.~Sutskever, J.~Bruna, D.~Erhan, I.~J. Goodfellow, and
  R.~Fergus, ``Intriguing properties of neural networks,'' in \emph{2nd
  International Conference on Learning Representations, {ICLR} 2014, Banff, AB,
  Canada, April 14-16, 2014, Conference Track Proceedings}, 2014. [Online].
  Available: \url{http://arxiv.org/abs/1312.6199}
\BIBentrySTDinterwordspacing

\bibitem{Sharif}
\BIBentryALTinterwordspacing
M.~Sharif, S.~Bhagavatula, L.~Bauer, and M.~K. Reiter, ``Accessorize to a
  crime: Real and stealthy attacks on state-of-the-art face recognition,'' in
  \emph{Proceedings of the 2016 {ACM} {SIGSAC} Conference on Computer and
  Communications Security, Vienna, Austria, October 24-28, 2016}, 2016, pp.
  1528--1540. [Online]. Available:
  \url{https://doi.org/10.1145/2976749.2978392}
\BIBentrySTDinterwordspacing

\bibitem{Lovisotto}
\BIBentryALTinterwordspacing
G.~Lovisotto, H.~Turner, I.~Sluganovic, M.~Strohmeier, and I.~Martinovic,
  ``{SLAP:} improving physical adversarial examples with short-lived
  adversarial perturbations,'' in \emph{30th {USENIX} Security Symposium,
  {USENIX} Security 2021, August 11-13, 2021}, 2021, pp. 1865--1882. [Online].
  Available:
  \url{https://www.usenix.org/conference/usenixsecurity21/presentation/lovisotto}
\BIBentrySTDinterwordspacing

\bibitem{RenH}
\BIBentryALTinterwordspacing
H.~Ren, T.~Huang, and H.~Yan, ``Adversarial examples: attacks and defenses in
  the physical world,'' \emph{Int. J. Mach. Learn. Cybern.}, vol.~12, no.~11,
  pp. 3325--3336, 2021. [Online]. Available:
  \url{https://doi.org/10.1007/s13042-020-01242-z}
\BIBentrySTDinterwordspacing

\bibitem{WangJ}
\BIBentryALTinterwordspacing
J.~Wang, ``Adversarial examples in physical world,'' in \emph{Proceedings of
  the Thirtieth International Joint Conference on Artificial Intelligence,
  {IJCAI} 2021, Virtual Event / Montreal, Canada, 19-27 August 2021}, 2021, pp.
  4925--4926. [Online]. Available:
  \url{https://doi.org/10.24963/ijcai.2021/694}
\BIBentrySTDinterwordspacing

\bibitem{Ding2020MMA}
\BIBentryALTinterwordspacing
G.~W. Ding, Y.~Sharma, K.~Y.~C. Lui, and R.~Huang, ``{MMA} training: Direct
  input space margin maximization through adversarial training,'' in \emph{8th
  International Conference on Learning Representations, {ICLR} 2020, Addis
  Ababa, Ethiopia, April 26-30, 2020}, 2020. [Online]. Available:
  \url{https://openreview.net/forum?id=HkeryxBtPB}
\BIBentrySTDinterwordspacing

\bibitem{HoJ}
\BIBentryALTinterwordspacing
J.~Ho, B.~Lee, and D.~Kang, ``Attack-less adversarial training for a robust
  adversarial defense,'' \emph{Appl. Intell.}, vol.~52, no.~4, pp. 4364--4381,
  2022. [Online]. Available: \url{https://doi.org/10.1007/s10489-021-02523-y}
\BIBentrySTDinterwordspacing

\bibitem{ZhangJ}
\BIBentryALTinterwordspacing
J.~Zhang, X.~Xu, B.~Han, G.~Niu, L.~Cui, M.~Sugiyama, and M.~S. Kankanhalli,
  ``Attacks which do not kill training make adversarial learning stronger,'' in
  \emph{Proceedings of the 37th International Conference on Machine Learning,
  {ICML} 2020, 13-18 July 2020, Virtual Event}, 2020, pp. 11\,278--11\,287.
  [Online]. Available: \url{http://proceedings.mlr.press/v119/zhang20z.html}
\BIBentrySTDinterwordspacing

\bibitem{Hyungyu}
\BIBentryALTinterwordspacing
H.~Lee, H.~Bae, and S.~Yoon, ``Gradient masking of label smoothing in
  adversarial robustness,'' \emph{{IEEE} Access}, vol.~9, pp. 6453--6464, 2021.
  [Online]. Available: \url{https://doi.org/10.1109/ACCESS.2020.3048120}
\BIBentrySTDinterwordspacing

\bibitem{LiY}
\BIBentryALTinterwordspacing
Y.~Li, S.~Cheng, H.~Su, and J.~Zhu, ``Defense against adversarial attacks via
  controlling gradient leaking on embedded manifolds,'' in \emph{Computer
  Vision - {ECCV} 2020 - 16th European Conference, Glasgow, UK, August 23-28,
  2020, Proceedings, Part {XXVIII}}, 2020, pp. 753--769. [Online]. Available:
  \url{https://doi.org/10.1007/978-3-030-58604-1\_45}
\BIBentrySTDinterwordspacing

\bibitem{AprilPyone}
\BIBentryALTinterwordspacing
M.~AprilPyone and H.~Kiya, ``Block-wise image transformation with secret key
  for adversarially robust defense,'' \emph{IEEE Transactions on Information
  Forensics and Security}, vol.~16, pp. 2709--2723, 2021. [Online]. Available:
  \url{https://doi.org/10.1109/TIFS.2021.3062977}
\BIBentrySTDinterwordspacing

\bibitem{Athalye}
\BIBentryALTinterwordspacing
A.~Athalye, N.~Carlini, and D.~A. Wagner, ``Obfuscated gradients give a false
  sense of security: Circumventing defenses to adversarial examples,'' in
  \emph{Proceedings of the 35th International Conference on Machine Learning,
  {ICML} 2018, Stockholmsm{\"{a}}ssan, Stockholm, Sweden, July 10-15, 2018},
  2018, pp. 274--283. [Online]. Available:
  \url{http://proceedings.mlr.press/v80/athalye18a.html}
\BIBentrySTDinterwordspacing

\bibitem{TanJ}
\BIBentryALTinterwordspacing
J.~Tan, N.~Ji, H.~Xie, and X.~Xiang, ``Legitimate adversarial patches: Evading
  human eyes and detection models in the physical world,'' in \emph{{MM} '21:
  {ACM} Multimedia Conference, Virtual Event, China, October 20 - 24, 2021},
  2021, pp. 5307--5315. [Online]. Available:
  \url{https://doi.org/10.1145/3474085.3475653}
\BIBentrySTDinterwordspacing

\bibitem{Tramer}
\BIBentryALTinterwordspacing
F.~Tramer, A.~Kurakin, N.~Papernot, I.~J. Goodfellow, D.~Boneh, and P.~D.
  McDaniel, ``Ensemble adversarial training: Attacks and defenses,'' in
  \emph{6th International Conference on Learning Representations, {ICLR} 2018,
  Vancouver, BC, Canada, April 30 - May 3, 2018, Conference Track Proceedings},
  2018. [Online]. Available: \url{https://openreview.net/forum?id=rkZvSe-RZ}
\BIBentrySTDinterwordspacing

\bibitem{BaiT}
\BIBentryALTinterwordspacing
T.~Bai, J.~Luo, J.~Zhao, B.~Wen, and Q.~Wang, ``Recent advances in adversarial
  training for adversarial robustness,'' in \emph{Proceedings of the Thirtieth
  International Joint Conference on Artificial Intelligence, {IJCAI} 2021,
  Virtual Event / Montreal, Canada, 19-27 August 2021}, 2021, pp. 4312--4321.
  [Online]. Available: \url{https://doi.org/10.24963/ijcai.2021/591}
\BIBentrySTDinterwordspacing

\bibitem{Lust}
\BIBentryALTinterwordspacing
J.~Lust and A.~P. Condurache, ``Efficient detection of adversarial,
  out-of-distribution and other misclassified samples,'' \emph{Neurocomputing},
  vol. 470, pp. 335--343, 2022. [Online]. Available:
  \url{https://doi.org/10.1016/j.neucom.2021.05.102}
\BIBentrySTDinterwordspacing

\bibitem{LuoW}
\BIBentryALTinterwordspacing
W.~Luo, C.~Wu, L.~Ni, N.~Zhou, and Z.~Zhang, ``Detecting adversarial examples
  by positive and negative representations,'' \emph{Appl. Soft Comput.}, vol.
  117, p. 108383, 2022. [Online]. Available:
  \url{https://doi.org/10.1016/j.asoc.2021.108383}
\BIBentrySTDinterwordspacing

\bibitem{KianiS}
\BIBentryALTinterwordspacing
S.~Kiani, S.~Awan, C.~Lan, F.~Li, and B.~Luo, ``Two souls in an adversarial
  image: Towards universal adversarial example detection using multi-view
  inconsistency,'' in \emph{{ACSAC} '21: Annual Computer Security Applications
  Conference, Virtual Event, USA, December 6 - 10, 2021}, 2021, pp. 31--44.
  [Online]. Available: \url{https://doi.org/10.1145/3485832.3485904}
\BIBentrySTDinterwordspacing

\bibitem{Aldahdooh}
\BIBentryALTinterwordspacing
A.~Aldahdooh, W.~Hamidouche, S.~A. Fezza, and O.~D{\'{e}}forges, ``Adversarial
  example detection for {DNN} models: {A} review,'' \emph{CoRR}, vol.
  abs/2105.00203, 2021. [Online]. Available:
  \url{https://arxiv.org/abs/2105.00203}
\BIBentrySTDinterwordspacing

\bibitem{Nesti}
\BIBentryALTinterwordspacing
F.~Nesti, A.~Biondi, and G.~C. Buttazzo, ``Detecting adversarial examples by
  input transformations, defense perturbations, and voting,'' \emph{CoRR}, vol.
  abs/2101.11466, 2021. [Online]. Available:
  \url{https://arxiv.org/abs/2101.11466}
\BIBentrySTDinterwordspacing

\bibitem{XuW}
\BIBentryALTinterwordspacing
W.~Xu, D.~Evans, and Y.~Qi, ``Feature squeezing: Detecting adversarial examples
  in deep neural networks,'' in \emph{25th Annual Network and Distributed
  System Security Symposium, {NDSS} 2018, San Diego, California, USA, February
  18-21, 2018}, 2018. [Online]. Available:
  \url{http://wp.internetsociety.org/ndss/wp-content/uploads/sites/25/2018/02/ndss2018\_03A-4\_Xu\_paper.pdf}
\BIBentrySTDinterwordspacing

\bibitem{LiuH}
\BIBentryALTinterwordspacing
H.~Liu, B.~Zhao, Y.~Peng, J.~Guo, and P.~Liu, ``Feature-filter: Detecting
  adversarial examples through filtering off recessive features,'' \emph{CoRR},
  vol. abs/2107.09502, 2021. [Online]. Available:
  \url{https://arxiv.org/abs/2107.09502}
\BIBentrySTDinterwordspacing

\bibitem{Graese}
\BIBentryALTinterwordspacing
A.~Graese, A.~Rozsa, and T.~E. Boult, ``Assessing threat of adversarial
  examples on deep neural networks,'' in \emph{15th {IEEE} International
  Conference on Machine Learning and Applications, {ICMLA} 2016, Anaheim, CA,
  USA, December 18-20, 2016}, 2016, pp. 69--74. [Online]. Available:
  \url{https://doi.org/10.1109/ICMLA.2016.0020}
\BIBentrySTDinterwordspacing

\bibitem{Bahat}
\BIBentryALTinterwordspacing
Y.~Bahat, M.~Irani, and G.~Shakhnarovich, ``Natural and adversarial error
  detection using invariance to image transformations,'' \emph{CoRR}, vol.
  abs/1902.00236, 2019. [Online]. Available:
  \url{http://arxiv.org/abs/1902.00236}
\BIBentrySTDinterwordspacing

\bibitem{TianS}
\BIBentryALTinterwordspacing
S.~Tian, G.~Yang, and Y.~Cai, ``Detecting adversarial examples through image
  transformation,'' in \emph{Proceedings of the Thirty-Second {AAAI} Conference
  on Artificial Intelligence, (AAAI-18), the 30th innovative Applications of
  Artificial Intelligence (IAAI-18), and the 8th {AAAI} Symposium on
  Educational Advances in Artificial Intelligence (EAAI-18), New Orleans,
  Louisiana, USA, February 2-7, 2018}, 2018, pp. 4139--4146. [Online].
  Available:
  \url{https://www.aaai.org/ocs/index.php/AAAI/AAAI18/paper/view/17408}
\BIBentrySTDinterwordspacing

\bibitem{Mekala}
\BIBentryALTinterwordspacing
R.~R. Mekala, G.~E. Magnusson, A.~Porter, M.~Lindvall, and M.~Diep,
  ``Metamorphic detection of adversarial examples in deep learning models with
  affine transformations,'' in \emph{Proceedings of the 4th International
  Workshop on Metamorphic Testing, MET@ICSE 2019, Montreal, QC, Canada, May 26,
  2019}, 2019, pp. 55--62. [Online]. Available:
  \url{https://doi.org/10.1109/MET.2019.00016}
\BIBentrySTDinterwordspacing

\bibitem{Guesmi}
\BIBentryALTinterwordspacing
A.~Guesmi, I.~Alouani, M.~Baklouti, T.~Frikha, and M.~Abid, ``{SIT:} stochastic
  input transformation to defend against adversarial attacks on deep neural
  networks,'' \emph{{IEEE} Des. Test}, vol.~39, no.~3, pp. 63--72, 2022.
  [Online]. Available: \url{https://doi.org/10.1109/MDAT.2021.3077542}
\BIBentrySTDinterwordspacing

\bibitem{ShouW}
W.~Shuo, N.~Surya, A.~Alsharif, R.~Carsten, and G.~Marthie, ``Adversarial
  detection by latent style transformations,'' \emph{IEEE Transactions on
  Information Forensics and Security}, vol.~17, pp. 1099--1114, 2022.

\bibitem{Kantaros}
\BIBentryALTinterwordspacing
Y.~Kantaros, T.~J. Carpenter, S.~Park, R.~Ivanov, S.~Jang, I.~Lee, and
  J.~Weimer, ``Visionguard: Runtime detection of adversarial inputs to
  perception systems,'' \emph{CoRR}, vol. abs/2002.09792, 2020. [Online].
  Available: \url{https://arxiv.org/abs/2002.09792}
\BIBentrySTDinterwordspacing

\bibitem{MagNet}
\BIBentryALTinterwordspacing
D.~Meng and H.~Chen, ``Magnet: {A} two-pronged defense against adversarial
  examples,'' in \emph{Proceedings of the 2017 {ACM} {SIGSAC} Conference on
  Computer and Communications Security, {CCS} 2017, Dallas, TX, USA, October 30
  - November 03, 2017}, 2017, pp. 135--147. [Online]. Available:
  \url{https://doi.org/10.1145/3133956.3134057}
\BIBentrySTDinterwordspacing

\bibitem{ZhouY}
\BIBentryALTinterwordspacing
Y.~Zhou, X.~Hu, J.~Han, L.~Wang, and S.~Duan, ``High frequency patterns play a
  key role in the generation of adversarial examples,'' \emph{Neurocomputing},
  vol. 459, pp. 131--141, 2021. [Online]. Available:
  \url{https://doi.org/10.1016/j.neucom.2021.06.078}
\BIBentrySTDinterwordspacing

\bibitem{Thang}
\BIBentryALTinterwordspacing
D.~D. Thang and T.~Matsui, ``Image transformation can make neural networks more
  robust against adversarial examples,'' \emph{CoRR}, vol. abs/1901.03037,
  2019. [Online]. Available: \url{http://arxiv.org/abs/1901.03037}
\BIBentrySTDinterwordspacing

\bibitem{ImageNet}
\BIBentryALTinterwordspacing
O.~Russakovsky, J.~Deng, H.~Su, J.~Krause, S.~Satheesh, S.~Ma, Z.~Huang,
  A.~Karpathy, A.~Khosla, M.~S. Bernstein, A.~C. Berg, and L.~Fei{-}Fei,
  ``Imagenet large scale visual recognition challenge,'' \emph{Int. J. Comput.
  Vis.}, vol. 115, no.~3, pp. 211--252, 2015. [Online]. Available:
  \url{https://doi.org/10.1007/s11263-015-0816-y}
\BIBentrySTDinterwordspacing

\bibitem{LiangB}
\BIBentryALTinterwordspacing
B.~Liang, H.~Li, M.~Su, X.~Li, W.~Shi, and X.~Wang, ``Detecting adversarial
  image examples in deep neural networks with adaptive noise reduction,''
  \emph{{IEEE} Trans. Dependable Secur. Comput.}, vol.~18, no.~1, pp. 72--85,
  2021. [Online]. Available: \url{https://doi.org/10.1109/TDSC.2018.2874243}
\BIBentrySTDinterwordspacing

\bibitem{Kherchouche}
\BIBentryALTinterwordspacing
A.~Kherchouche, S.~A. Fezza, and W.~Hamidouche, ``Detect and defense against
  adversarial examples in deep learning using natural scene statistics and
  adaptive denoising,'' \emph{CoRR}, vol. abs/2107.05780, 2021. [Online].
  Available: \url{https://arxiv.org/abs/2107.05780}
\BIBentrySTDinterwordspacing

\bibitem{Sundararajana}
\BIBentryALTinterwordspacing
M.~Sundararajan, A.~Taly, and Q.~Yan, ``Axiomatic attribution for deep
  networks,'' in \emph{Proceedings of the 34th International Conference on
  Machine Learning, {ICML} 2017, Sydney, NSW, Australia, 6-11 August 2017},
  2017, pp. 3319--3328. [Online]. Available:
  \url{http://proceedings.mlr.press/v70/sundararajan17a.html}
\BIBentrySTDinterwordspacing

\bibitem{Sundararajanb}
\BIBentryALTinterwordspacing
P.~K. Mudrakarta, A.~Taly, M.~Sundararajan, and K.~Dhamdhere, ``Did the model
  understand the question?'' in \emph{Proceedings of the 56th Annual Meeting of
  the Association for Computational Linguistics, {ACL} 2018, Melbourne,
  Australia, July 15-20, 2018, Volume 1: Long Papers}, 2018, pp. 1896--1906.
  [Online]. Available: \url{https://aclanthology.org/P18-1176/}
\BIBentrySTDinterwordspacing

\bibitem{shrikumar}
\BIBentryALTinterwordspacing
A.~Shrikumar, P.~Greenside, and A.~Kundaje, ``Learning important features
  through propagating activation differences,'' in \emph{Proceedings of the
  34th International Conference on Machine Learning, {ICML} 2017, Sydney, NSW,
  Australia, 6-11 August 2017}, 2017, pp. 3145--3153. [Online]. Available:
  \url{http://proceedings.mlr.press/v70/shrikumar17a.html}
\BIBentrySTDinterwordspacing

\bibitem{Tomsett}
\BIBentryALTinterwordspacing
R.~Tomsett, D.~Harborne, S.~Chakraborty, P.~Gurram, and A.~D. Preece, ``Sanity
  checks for saliency metrics,'' in \emph{The Thirty-Fourth {AAAI} Conference
  on Artificial Intelligence, {AAAI} 2020, The Thirty-Second Innovative
  Applications of Artificial Intelligence Conference, {IAAI} 2020, The Tenth
  {AAAI} Symposium on Educational Advances in Artificial Intelligence, {EAAI}
  2020, New York, NY, USA, February 7-12, 2020}, 2020, pp. 6021--6029.
  [Online]. Available:
  \url{https://ojs.aaai.org/index.php/AAAI/article/view/6064}
\BIBentrySTDinterwordspacing

\bibitem{Qudrat}
\BIBentryALTinterwordspacing
Q.~E.~A. Ratul, E.~Serra, and A.~Cuzzocrea, ``Evaluating attribution methods in
  machine learning interpretability,'' in \emph{2021 {IEEE} International
  Conference on Big Data (Big Data), Orlando, FL, USA, December 15-18, 2021},
  2021, pp. 5239--5245. [Online]. Available:
  \url{https://doi.org/10.1109/BigData52589.2021.9671501}
\BIBentrySTDinterwordspacing

\end{thebibliography}

\end{document}